\documentclass[final,5p,times,twocolumn]{elsarticle}

\usepackage{utils}

\usepackage{xcolor}
\usepackage{amsmath} %
\usepackage{siunitx}
\usepackage{cuted}
\usepackage{fancyhdr}

\journal{Review}

\begin{document}
\let\today\relax
\makeatletter
\def\ps@pprintTitle{%
    \let\@oddhead\@empty
    \let\@evenhead\@empty
    \def\@oddfoot{\footnotesize\itshape
         {Submitted preprint} \hfill\today}%
    \let\@evenfoot\@oddfoot
    }
\makeatother

\begin{frontmatter}

\title{Beyond Role-Based Surgical Domain Modeling: \\ Generalizable Re-Identification in the Operating Room}

\author[1]{
    Tony Danjun Wang 
    \fnref{fn1}
}
\ead{tony.wang@tum.de}

\author[1]{
    Lennart Bastian
    \fnref{fn1}%
    \corref{cor1}
} 
\ead{lennart.bastian@tum.de}

\author[2]{
    Tobias Czempiel
} 
\ead{t.czempiel@ucl.ac.uk }

\author[3]{
    Christian Heiliger%
} 
\ead{christian.heiliger@med.uni-muenchen.de}

\author[1]{
    Nassir Navab%
} 
\ead{nassir.navab@tum.de}

\cortext[cor1]{Corresponding author}
\fntext[fn1]{Equal Contributions} 

\affiliation[1]{organization={Chair for Computer Aided Medical Procedures, Technical University of Munich}, 
    addressline={Boltzmannstraße 3},
    postcode={85748}, 
    city={Garching}, 
    country={Germany}}
    
\affiliation[2]{organization={UCL Hawkes Institute, Dept. Computer Science, University College London}, 
    addressline={43-45 Foley St}, 
    city={London},
    postcode={W1W 7TY},
    country={United Kingdom}}

\affiliation[3]{organization={Department of General, Visceral, and Transplant Surgery, University Hospital, Ludwig-Maximilians-University},
    addressline={Marchioninistr. 15}, 
    city={Munich},
    postcode={81377},
    country={Germany}}

\begin{abstract}
Surgical domain models improve workflow optimization through automated predictions of each staff member's surgical role.
However, mounting evidence indicates that team familiarity and individuality impact surgical outcomes.
We present a novel staff-centric modeling approach that characterizes individual team members through their distinctive movement patterns and physical characteristics, enabling long-term tracking and analysis of surgical personnel across multiple procedures.
To address the challenge of inter-clinic variability, we develop a generalizable re-identification framework that encodes sequences of 3D point clouds to capture shape and articulated motion patterns unique to each individual. 
Our method achieves 86.19\% accuracy on realistic clinical data while maintaining 75.27\% accuracy when transferring between different environments -- a 12\% improvement over existing methods. 
When used to augment markerless personnel tracking, our approach improves accuracy by over 50\%.
Through extensive validation across three datasets and the introduction of a novel workflow visualization technique, we demonstrate how our framework can reveal novel insights into surgical team dynamics and space utilization patterns, advancing methods to analyze surgical workflows and team coordination.
\end{abstract}


\end{frontmatter}

\section{Introduction}
\label{sec:introduction}

\begin{figure}[ht]
    \centering
    \includegraphics[width=\columnwidth]{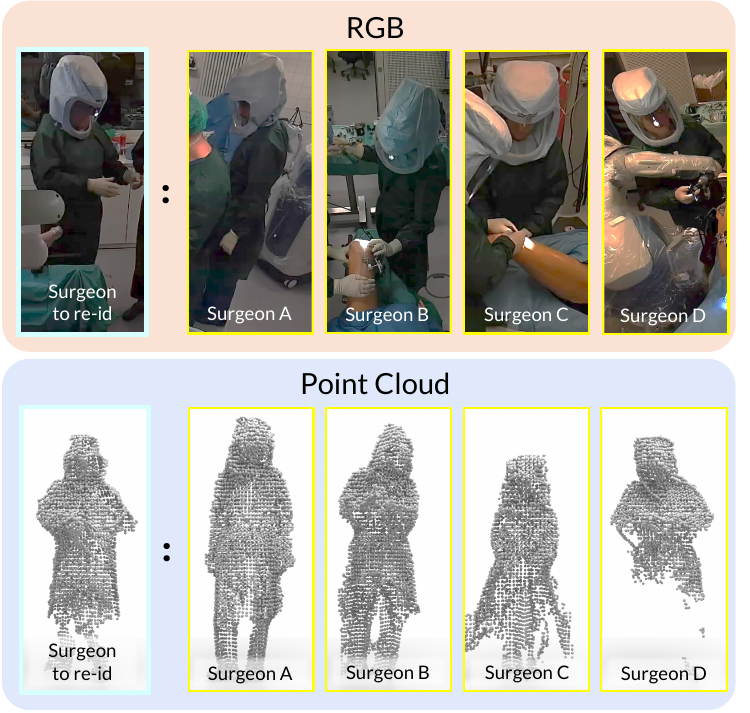}
    \caption{
    \textit{Who is the surgeon to re-identify?}
    A depiction of four surgeons (persons A, B, C, and D), which are challenging to visually differentiate.
    Existing re-identification methods relying on \colorbox{rgbexp!15}{RGB (top)} images often depend on facial features, attire, complexion, or physique. 
    These distinctive features are obscured in ORs, providing an obstacle to such methods;
    we observe that encoding \colorbox{depthexp!15}{3D point clouds (bottom)} proves more effective. 
    By encapsulating characteristics such as an individual's stature, body shape, proportions, and volume, methods can more easily differentiate individuals than their clinical role.
    For example, these four surgeons can be more easily distinguished based on their height differences, which is ambiguous in RGB images due to the loss of absolute scale.
    \textit{The surgeon to re-identify is Person D.}
    }
    \label{fig:teaser}
\end{figure}

In dynamic and high-stakes surgical operating rooms (OR), subtle decisions can profoundly influence patient outcomes.
Surgical Data Science (SDS) has emerged as a pioneering field in medical research, empowering surgical teams to enhance their coordination, minimize inefficiencies, and ultimately elevate the standard for patient care \cite{maier2022surgical}.
A primary objective of SDS is OR workflow analysis: examining clinicians' movements, roles, and interactions within and around the OR. 
This encompasses tasks like phase and action recognition \cite{sharghi2020automatic,czempiel2022surgical,garrow2021machine,bastian2023know}, person anonymization \cite{flouty2018faceoff,bastian2023disguisor}, as well as human pose recognition \cite{belagiannis2016parsing,srivastav2018mvor,gerats20233d,liu2024human}, object and instance segmentation \cite{li2020robotic,bastian2023segmentor}, and role-prediction \cite{ozsoy20224d}.
These methodological developments serve as building blocks for intelligent systems that are already complementing or exceeding human performance in specific tasks \cite{Varghese2024}, from providing real-time surgical team support \cite{Varghese2024,Moulla2020} to wearable nurse assistance \cite{cramer2024requirement}, personalized feedback, \cite{laca_UsingRealtimeFeedback_2022} and automated report generation \cite{lin_SGTSceneGraphGuided_2022,xu_ClassIncrementalDomainAdaptation_2021}.

Each role in the OR has unique responsibilities primarily orchestrated by the surgeon, who coordinates team activities and maintains situational awareness across the surgical workflow in addition to performing the operative procedure.
Studies suggest that this coordinated effort increases cognitive load; intelligent systems have the potential to guide each role towards harmonized interactions in the OR, alleviating a cognitive burden that impacts error rates and performance \cite{Avrunin2018, CSTReview2021}. 

The existing literature focuses extensively on a surgeon's skill \cite{zia2018automated,levin2019automated,liu2021towards} while giving less attention to the dynamic interactions and collaborative processes among surgical team members.
Furthermore, current learning-based approaches to OR analysis and optimization often treat surgical staff as interchangeable components within their designated roles. 
This perspective, while useful for coarse workflow analysis, fails to account for the nuanced differences between surgeons, assistants, nurses, and other team members. 
Lack of team familiarity and poor teamwork coordination are associated with increased operative times, higher complication rates, and elevated costs \cite{mazzocco_SurgicalTeamBehaviors_2009,al2023artificial,pasquer_OperatingRoomOrganization_2024}. 
Conversely, stable and well-synchronized teams demonstrate improved performance across various metrics, including shorter operative durations, lower morbidity rates, and reduced length of hospital stay \cite{pasquer_OperatingRoomOrganization_2024}.
Intelligent systems in the OR should, therefore, be aware of each individual's role in the greater coordinated effort of surgery and maintain a nuanced understanding of how the workflow changes for each team constellation.

\subsection{A Staff-centric Approach}
A concrete example that underscores the need for \textit{staff-centric} approaches is the varying skill levels of surgical staff.
Role-based modeling cannot differentiate between a novice resident and an experienced attending surgeon when providing workflow guidance or suggestions. 
However, the needs, behaviors, and impact on surgical outcomes can differ significantly between individuals and play a prominent role in the team dynamics~\cite{parker2020impact,laca_UsingRealtimeFeedback_2022,petrosoniak_TrackingWorkflowHighstakes_2018}.
Additional aspects, for instance, the complexity introduced by new technological equipment like multi-cart robotic systems \cite{verma2021multi}, also cause team interactions to change in unforeseeable ways, further underscoring the need to look beyond role-based modeling.

While the SDS literature has previously addressed OR personnel tracking \cite{belagiannis2016parsing,huMultiCameraMultiPersonTracking2022a} and role prediction \cite{ozsoy20224d}, long-term tracking and re-identification remain unexplored. 
Current approaches cannot establish correspondences over extended timescales - days, weeks, or even years - which are essential for understanding team dynamics and skill development. 
In contrast to human-pose-based tracking methods, re-identification could enable fine-grained workflow analysis across multiple surgeries without requiring manual intervention or disruptive physical tracking devices.
To address these challenges, we propose to model OR personnel through the lens of person re-identification.
This paradigm not only enables longitudinal surgical analysis but could also serve as a means for making human-pose trackers robust to track-swapping and missed detections, a problem in crowded OR environments (see \autoref{subsec:results_human_pose_based_tracking}).
To date, no markerless approach is capable of drawing long-term associations for OR personnel.

\begin{figure*}[ht]
    \centering
    \includegraphics[width=\textwidth]{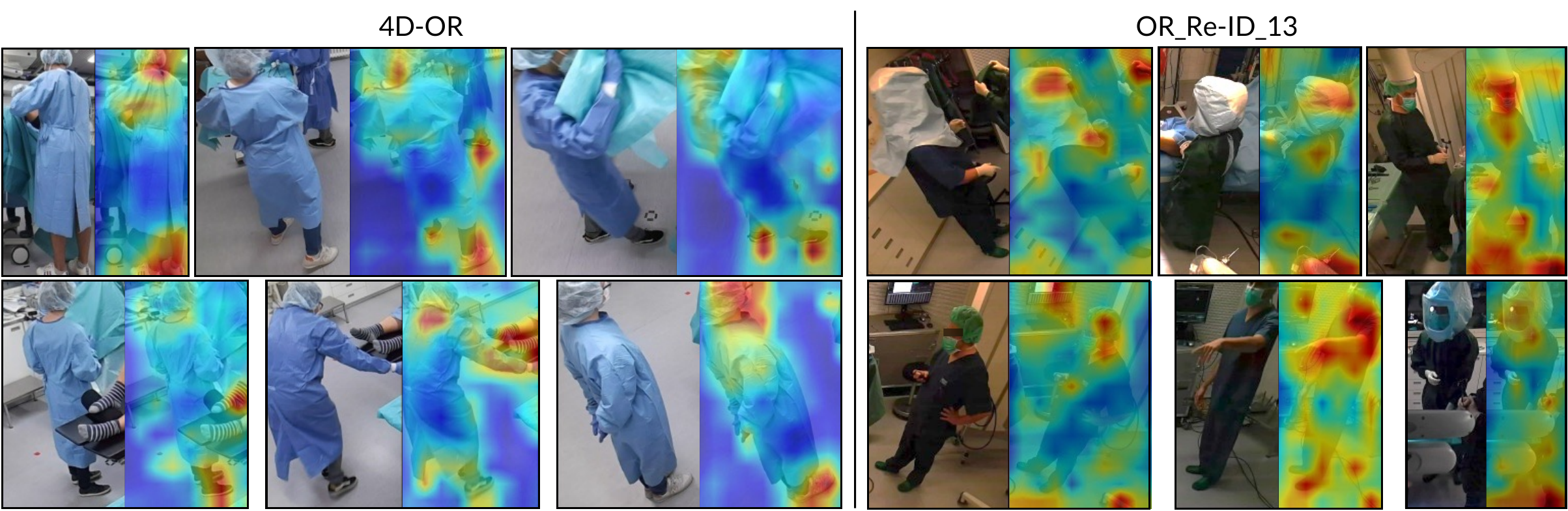}
    \caption{RGB image excepts and overlayed saliency maps generated with GradCAM \cite{selvaraju2017grad} on the simulated datasets 4D-OR \cite{ozsoy20224d} and \ordataset.
    4D-OR's limited realism and variety allow CNNs to identify individuals solely by their heads and shoes. In more realistic OR settings like \ordataset, these features become less useful due to more homogenous attire. These discrepancies between different clinical environments can impede generalization.}
    \label{fig:saliency_maps}
\end{figure*}

\subsection{Challenges in Surgical Operating Rooms}

The OR presents unique challenges distinct from the general computer vision domain, which require deliberate modeling.
Conventional RGB-based re-id methods often focus on the head, shoulders, and feet \cite{liInDepthExplorationPerson2023}. 
In surgical ORs, standardized attire such as smocks, face masks, and skull caps obscure these key identifying features, rendering traditional approaches ineffective (see \autoref{fig:teaser}).

Moreover, deep neural networks exhibit a bias towards texture over shape \cite{geirhosImageNettrainedCNNsAre2022}, posing challenges as attire is homogeneous within but varies significantly between clinics (see \autoref{fig:saliency_maps}) \cite{liu2024human}.
We hypothesize that this bias can be overcome by emphasizing individuals' distinctive shape and motion rather than texture indicative of a specific clinic or role in the surgical team.

\subsection{Generalizable Re-identification for the OR}

Based on these observations, we propose a novel approach to track and identify OR personnel by analyzing biometrics, such as body shape and articulated motion patterns, which remain distinguishable between individuals even when their attire is identical.
Our method encodes sequences of 3D point clouds for each individual, which are segmented from a global 4D representation of the OR.
This approach mitigates the reliance on appearance or role-based interactions, emphasizing shape and motion cues that distinguish an individual.
Our method demonstrates significantly improved generalization across diverse scenarios, paving the way for more adaptive and personalized intelligent systems in surgical settings.

\begin{figure*}[ht]
    \centering
    \includegraphics[width=\textwidth]{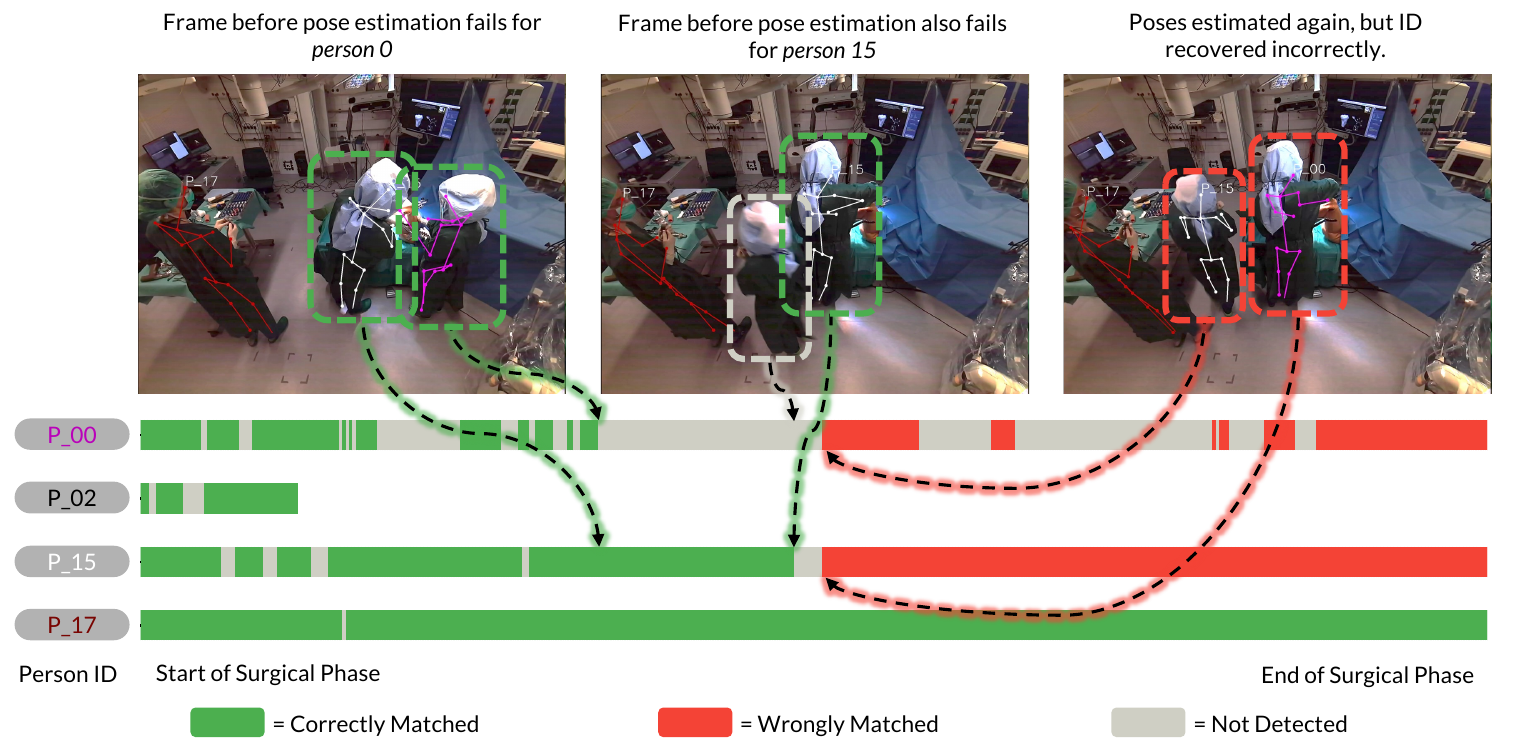}
    \caption{Person tracking using a heuristic-based approach by associating 3D human key points over time. 
    Each bar represents the timeline of an individual during the surgical phase. 
    Human pose estimation \cite{Srivastav_2024_CVPR} fails for \textit{person 0} just as they move to the left (from the depicted camera angle) of \textit{person 15}, and after some frames also for \textit{person 15}.
    When the 3D human pose estimator resumes detection in the subsequent frames, a heuristic-based tracking method \cite{ozsoy20224d} incorrectly associates the poses of \textit{persons 0} and \textit{15}, due to a missed detection.
    After a wrong association, tracks can be irreversibly mismatched for the remainder of a surgery.}
    \label{fig:joints_3d_figure}
\end{figure*}

Our main contributions can be summarized as follows:

\begin{itemize}
    \item We introduce the concept of staff-centric surgical domain modeling, emphasizing the importance of individual characteristics and team dynamics in surgical environments. This approach enables staff to be associated over long periods, facilitating analysis of team dynamics.
    \item We present a comprehensive analysis of the challenges in modeling OR personnel, highlighting the limitations of existing tracking methods and demonstrating why conventional texture-based re-identification techniques underperform in the OR or exhibit bias.
    \item We propose a novel approach for modeling OR personnel by implicitly abstracting shape and articulated motion cues from 3D point cloud sequences. This method addresses OR-specific challenges and demonstrates superior generalization across different OR environments, differentiating individuals rather than just surgical roles.
    \item We validate our approach through cross-evaluation on OR-specific datasets and more general outdoor person re-identification scenarios, illustrating its robustness and versatility in various settings.
    \item We showcase the practical application of our method by generating role-specific \textit{3D activity imprints}, offering insights into OR team dynamics and individual movement patterns. This visualization technique provides a foundation for data-driven optimization of surgical processes, potentially improving efficiency, training, and patient care.
\end{itemize}

\section{Related Works}
\label{sec:related_works}

\subsection{Operating Room Personnel Tracking}

\subsubsection{Markered and Manual Tracking}

Due to the lack of reliable automated personnel tracking solutions, some OR workflow studies have relied on manual inspections conducted by observers present in the OR \cite{birgandAttitudesRiskInfection2014,keller_DisruptiveBehaviorOperating_2019,petrosoniak_TrackingWorkflowHighstakes_2018,weston_UsingObservationBetter_2022}, manual annotations on video material \cite{petrosoniak_TrackingWorkflowHighstakes_2018,joseph_MinorFlowDisruptions_2019} or physical marker-based tracking systems \cite{birgandMotioncaptureSystemAssess2019,vankipuram_AutomatedWorkflowAnalysis_2011} to obtain staff identities and movement patterns. 
These approaches, however, present several drawbacks. 
Manual tracking is cumbersome and error-prone, while marker-based motion capture systems, such as those utilizing motion-capture suits\footnote{An example is the Vicon system \cite{vicon2024}, which uses multiple cameras to track reflective markers attached to subjects for precise motion capture.}, require additional hardware like reflective trackers \cite{birgandMotioncaptureSystemAssess2019,azevedo-costeTrackingClinicalStaff2019}. 
Alternatives such as RFID~\cite{rfid,vankipuram_AutomatedWorkflowAnalysis_2011} or Zigbee~\cite{bluetooth} similarly require supplementary hardware to assign unique IDs to individuals.

Systems requiring physical tags or reflective markers can interfere with sterilization and ergonomic considerations, potentially restricting movement or compromising the validity of workflow studies. 
These systems are also prone to noise \cite{haque2017towards,petrosoniak_TrackingWorkflowHighstakes_2018} and can introduce additional points of failure.
They can be particularly problematic in settings with heightened infection risk \cite{petrosoniak_TrackingWorkflowHighstakes_2018}, posing sterility concerns that complicate hygiene protocols \cite{birgandAttitudesRiskInfection2014}.
Furthermore, motion capture suits may exacerbate the Hawthorne effect\footnote{The Hawthorne effect refers to the tendency of people to change their behavior when they know they are being observed, potentially leading to skewed research results \cite{mccambridge_SystematicReviewHawthorne_2014}}, altering staff behavior due to awareness of being monitored.

Consequently, unobtrusive ``fly-on-the-wall" \footnote{"Fly-on-the-wall" refers to an observational technique where the observer (or in this case, the tracking system) is as unobtrusive and non-interfering as possible, as if it were a fly on the wall, allowing subjects to behave naturally without awareness of being monitored \cite{weston_UsingObservationBetter_2022}} style tracking systems are particularly compelling. 
These systems require no modification to the surgical workflow and are impervious to physical interference or failures associated with wearable devices, offering a promising solution for accurate and non-disruptive personnel tracking in OR environments.

\subsubsection{Markerless Human Pose-based Tracking}
While purely visual-based tracking systems have been used for personnel tracking \cite{belagiannis2016parsing,huMultiCameraMultiPersonTracking2022a,ozsoy20224d,liu2024human}, they typically rely on frame-by-frame human-pose estimation, upon which trajectories are estimated and clustered into \textit{tracklets} based on the proximity of subsequent pose estimates.
One unequivocal point of failure of such systems is that they cannot re-associate individuals who leave and re-enter the operating room, a common occurrence for circulating nurses or specialists when deviations from the standard workflow occur or additional equipment is needed from outside the OR \cite{neyens_UsingSystemsApproach_2019}.
However, the many obstructions in the OR can impact detection accuracy; given that these systems require an uninterrupted flow of information, an accumulation of errors inevitably yields erroneous tracks later in a given surgery (illustrated in \autoref{fig:joints_3d_figure}).

\subsection{Person Re-Identification.}

\begin{figure*}[htb]
    \centering
    \includegraphics[width=\textwidth]{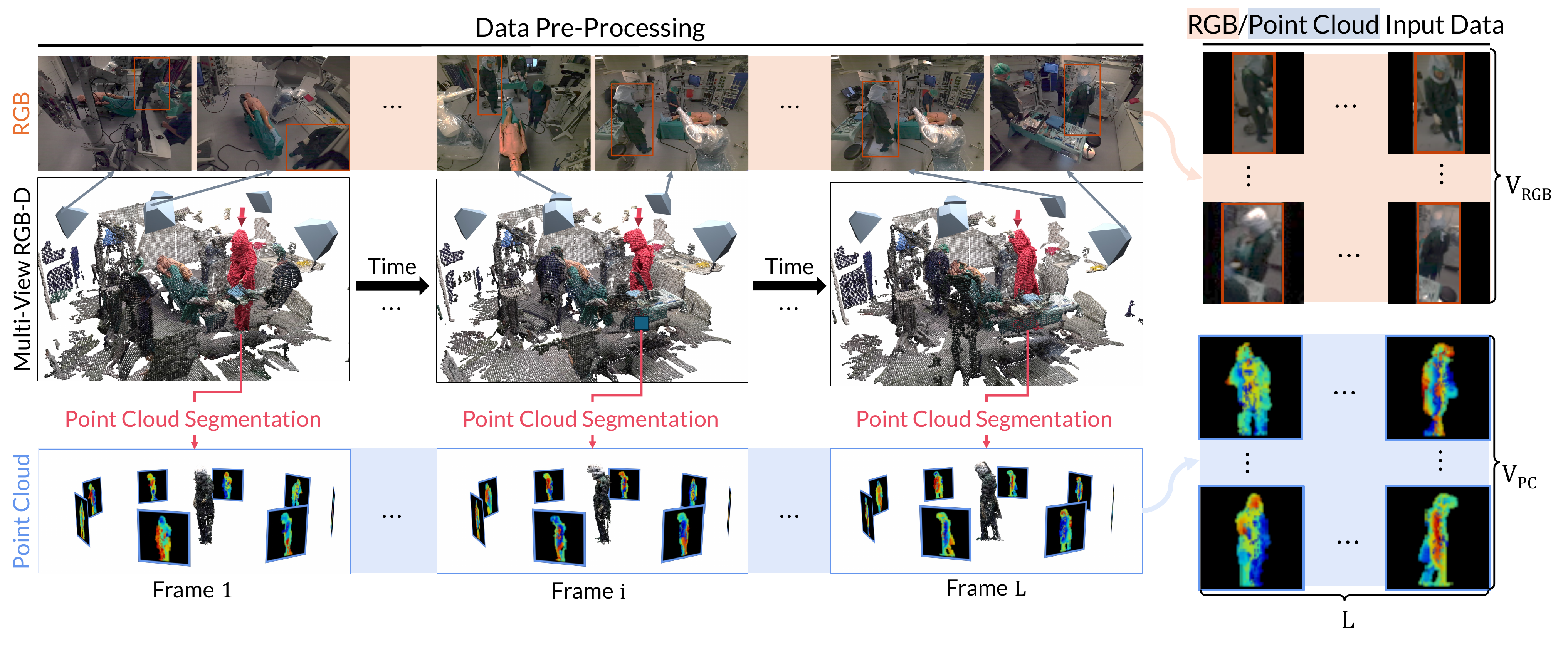}
    \caption{
    An overview of the data pre-processing to acquire RGB and point cloud input frames.
    From the raw multi-view RGB-D recordings, we first segment each individual from the fused 3D point cloud sequence using weakly-supervised 3D point cloud segmentation \cite{bastian2023segmentor}. 
    The resulting person point cloud is then used to render 2D depth maps using virtual rendering and to acquire the person bounding boxes to extract the 2D bounding boxes from the RGB images.
    Note that this example depicts the process for a single person. In practice, we repeat the same process for each person in the scene.
    }
    \label{fig:main-figure}
\end{figure*}

Person re-identification (re-id) has been explored extensively in the computer vision literature \cite{ming2022deep,zahra2023person,reid_review}.
Ming et al. \cite{ming2022deep} propose a 4-class taxonomy for re-identification methods into deep metric learning, local feature learning, GAN-based feature learning, and sequence feature learning.
However, the OR poses unique challenges for feature-based methods, as uniform attire such as scrubs, shoes, and masks leaves few reliable identifying features (see \autoref{fig:saliency_maps}). 
Instead of relying on unindicative RGB features, contrastive metric learning poses a convenient paradigm for encoding person sequences in a modality-independent way.

In contrast to classification-based approaches, contrastive or triplet-based losses enable scaling re-id models beyond known identities or those seen in a training set ~\cite{leibe_triplet}.
Instead, after training a discriminative similarity function,  a \textit{gallery} of known identities is constructed and compared to a \textit{probe} set of unknown identities. 
For a given probe, the closest match in the gallery is thus identified as the subject of interest.

\subsubsection{Clothes-Changing Person Re-Identification.}

In the computer vision literature, several works have addressed cloth-changing person re-identification (CC-re-ID) \cite{han_ClothingChangeFeatureAugmentation_, jin_ClothChangingPersonReidentification_2022, xu_DeepChangeLongTermPerson_2023, yang_GoodBadCausality_2023a} and long-term person re-identification (LT-re-ID) \cite{liInDepthExplorationPerson2023,liuLearningClothingPose2023,huang_ClothingStatusAwareness_2021}. 
These works aim to re-identify people invariant to the clothes they wear, similar to how one may want to extract unique biometrics in the OR despite similar attire. 
\cite{yang_GoodBadCausality_2023a} introduces a causality-based auto-intervention model to reduce the clothing bias for CC-reID, however, it requires the perception of clothing as a prerequisite, which is limited by clothing annotations and sample numbers.
Other works explore the disentanglement of shape and appearance for LT-reID \cite{liInDepthExplorationPerson2023,liuLearningClothingPose2023}.
However, these methods rely heavily on the ability to accurately separate the modeling of body shape from garments, which poses significant additional challenges in OR settings \cite{srivastav2018mvor,bastian2023disguisor,gerats20233d} due to baggy OR scrubs, the absence of explicit cloth models and frequent occlusions.

\subsubsection{Domain Generalizable Person Re-Identification.}

Traditional re-id systems often struggle with domain shifts, as models trained in one environment perform poorly in another due to variations in lighting, clothing, or sensor data.
To address these limitations, domain generalizable re-identification (DG-re-ID) methods aim to enhance the transferability of re-id models across different domains \cite{choi_MetaBatchInstanceNormalization_2021, dai_GeneralizablePersonReidentification_2021, dou_IdentitySeekingSelfSupervisedRepresentation_2023,jin_StyleNormalizationRestitution_2020, li_MitigateDomainShift_2024, ni_MetaDistributionAlignment_2022, ni2023part,song_GeneralizablePersonReIdentification_2019, xu_MimicEmbeddingAdaptive_2022}.
For instance, Ni et al. \cite{ni2023part} proposes a Part-Aware-Transformer (PAT) that employs a proxy task, mining local similarities between different identities and a part-guided self-distillation module to capture both local and global features, even when dealing with occlusions and changes in clothing. 
Such characteristics are particularly compelling in OR environments where occlusions and large domain shifts between different clinics pose significant challenges.
On the other hand, Song et al. \cite{song_GeneralizablePersonReIdentification_2019} attempt to develop domain-invariant representations within a meta-learning framework. 
Additionally, Ni et al. \cite{ni_MetaDistributionAlignment_2022} align source and target feature distributions with a prior distribution to achieve better generalization across domains.
In contrast to these approaches, our work aims to enhance domain generalization by leveraging dynamic 3D point cloud sequences instead of the more traditional RGB videos with the intuition that shape and motion characterize individuals regardless of significant domain shifts in the surrounding clinical environment. 
By doing so, we seek to minimize inherent domain differences and achieve a higher level of generalization.

\subsubsection{Re-identification through Biometrics and Gait}

Another branch of work focuses on leveraging biometrics, such as an individual's gait pattern, to learn cloth agnostic representations for robust re-id 
\cite{ahn2VGaitGaitRecognition2022, dou_GaitGCIGenerativeCounterfactual_2023, fanGaitPartTemporalPartBased2020, fanOpenGaitRevisitingGait2023, GaitbasedPersonIdentification, zhengGaitRecognitionWild2022, harrisSurveyHumanGaitBased2022, kwonComparativeStudyMarkerless2021, liaoModelbasedGaitRecognition2020, lidargait, teepe_DeeperUnderstandingSkeletonbased_2022, teepeGaitGraphGraphConvolutional2021, wangCombiningSilhouetteSkeleton2023, zhengGaitRecognitionWild2022, zhuEcReIDEnhancingCorrelations2023}.
These methods aim to re-identify individuals by taking sequences of a person's gait in the form of silhouettes 
\cite{dou_GaitGCIGenerativeCounterfactual_2023, fanGaitPartTemporalPartBased2020, fanOpenGaitRevisitingGait2023,wangCombiningSilhouetteSkeleton2023}, human pose key-points \cite{kwonComparativeStudyMarkerless2021,liaoModelbasedGaitRecognition2020,teepe_DeeperUnderstandingSkeletonbased_2022,teepeGaitGraphGraphConvolutional2021,wangCombiningSilhouetteSkeleton2023,zhuEcReIDEnhancingCorrelations2023}, or depth information \cite{ahn2VGaitGaitRecognition2022,GaitbasedPersonIdentification,zhengGaitRecognitionWild2022,lidargait}
as input.
When considering how such methods would be useful in the OR, one must consider numerous differences from typical street-wear re-identification scenarios, as in \cite{lidargait}.
Individuals typically do not walk long distances but are relatively stationary, challenging the practicality of gait analysis.
Furthermore, in contrast to street-wear datasets where subjects are in plain view of a camera positioned to capture a side profile of the viewer, OR datasets are crowded, and acquisitions contain severe occlusions \cite{bastian2023disguisor,liu2024human}.
Aggregated multi-view inputs can, therefore, exhibit greater partiality or noise, as occlusions or different camera setups often limit information visible from each sensor (see \autoref{fig:teaser}).
Furthermore, methods are not designed to leverage the long-form dependencies surgeries take the form of, rather focused on re-identifying individuals based on gait within short intervals or between different cameras on a street.

We incorporate these nuances into our modeling decisions and demonstrate that person segmentations from crowded 4D scene representations obtained by a weakly supervised 3D segmentation method suffice to discriminate individuals through their body shape and articulated motion over time.
Given these insights, we design a general framework to study the challenges of re-identification and human motion modeling in the OR.
We show that shape and motion representations are more generalizable than those obtained from RGB textures.
These findings are significant for multiple challenges in surgical workflow analysis, where a fundamental building block is both the tracking and re-identification of personnel despite context changes, occlusions, and movement constraints.

\subsection{Operating Room Workflow Visualizations}

OR environments are dynamic and increasingly feature varying digital devices.
Integrating new equipment typically requires adapting the surgical workflow; intelligent systems must be able to reckon with these changes \cite{wong2023workflow}.
Various visualizations, such as workflow heatmaps, have been employed to analyze OR workflows, to, for instance, provide insights into the efficiency of individuals' movement \cite{birgandMotioncaptureSystemAssess2019,liu2024human,mousavi_ObservationalStudyDoor_2018,petrosoniak_TrackingWorkflowHighstakes_2018}.
Of importance is whether new equipment causes workflow disruptions \cite{wong2023workflow}, line of sight issues to digital displays for endoscopic or laparoscopic cameras, or identifying areas where additional training or adjustments may be required.

Previous heatmap approaches include plotting staff movement on different custom backgrounds.
A straightforward approach involves plotting accumulated movements on a 2D canvas \cite{birgandMotioncaptureSystemAssess2019,liu2024human}.
However, this abstract representation lacks spatial context, making movement interpretation with respect to the spatial OR context challenging.
In contrast, 2D OR floor plans provide better spatial orientation \cite{mousavi_ObservationalStudyDoor_2018} but require manually creating floor plans that accurately represent each procedure.
Another approach overlays the heatmaps on the raw RGB images \cite{petrosoniak_TrackingWorkflowHighstakes_2018}.
While this approach preserves the intuitive visual context of the OR, the depth and scale ambiguity of 2D RGB images drastically limits the interpretations of the visualization.
In contrast to these methods, we propose \textit{3D activity imprints}, which plot OR personnel movements with respect to our 3D representation of the scene.

\begin{figure*}[ht]
    \centering
    \includegraphics[width=\textwidth]{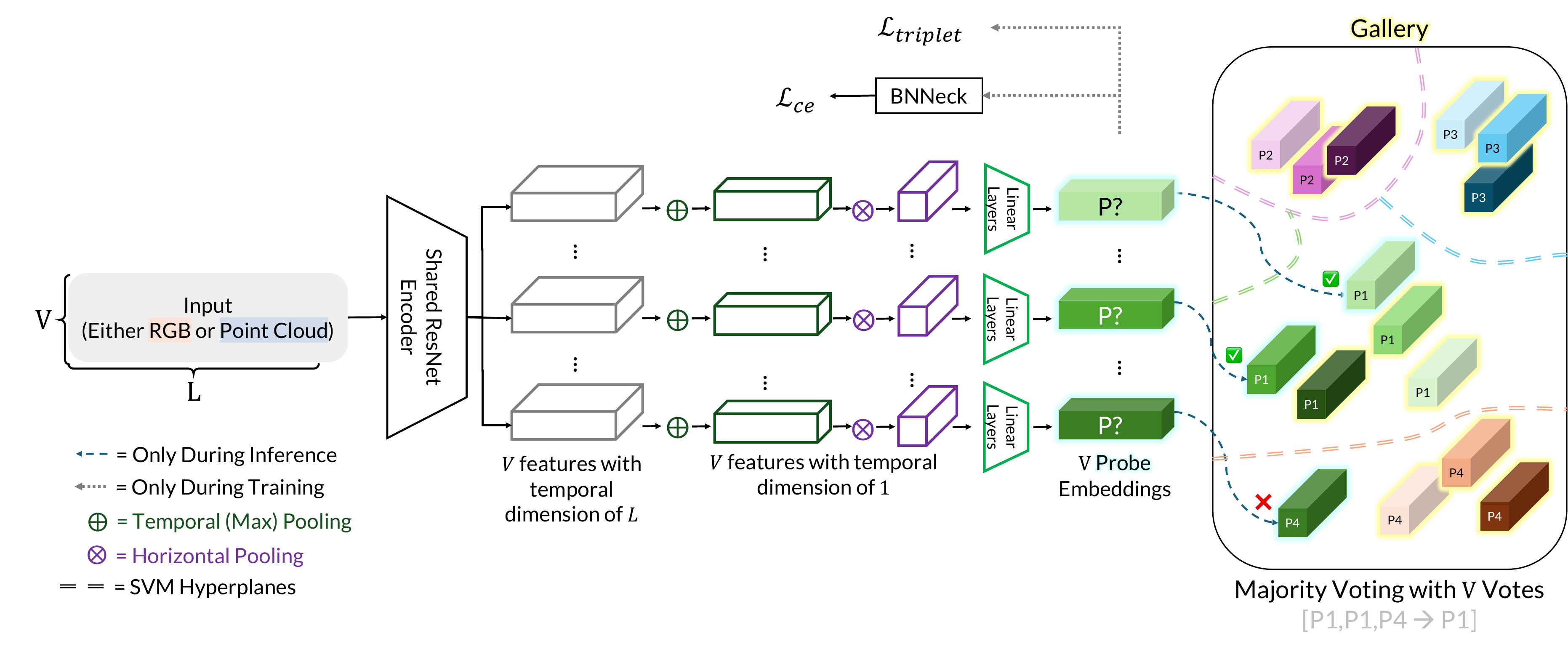}
    \caption{
    Our model can take a sequence of multi-view 2D images as input.
    These sequences can be RGB images or rendered point cloud images.
    Each image is individually encoded using a shared lightweight ResNet-9 encoder. Subsequently, each view is processed independently and trained with a combination of contrastive loss and cross-entropy loss.
    During inference, we use the gallery features to train an SVM, which separates the latent space into hyperplanes. 
    Each probe embedding is then assigned to a cluster, and we apply majority voting across the views to obtain our final prediction.
    }
    \label{fig:main-figure_part_2}
\end{figure*}

\subsection{OR Workflow Datasets}

While several OR workflow datasets are publicly available \cite{srivastav2018mvor,ozsoy20224d,belagiannis2016parsing}, they typically lack necessary annotations, such as person segmentation and identification labels, and involve only a limited number of individuals.
Belagiannis et al. \cite{belagiannis2016parsing} does not include participant identification labels, while MVOR \cite{srivastav2018mvor} lacks participant identification labels and consistent frame rates.
We, therefore, focus our analysis on the publically available 4D-OR \cite{ozsoy20224d}, as well as an internally acquired dataset \ordataset which contains longer acquisitions with more individuals (see \autoref{subsubsec:datasets}).

\section{Methods}
\label{section:methods}

\subsection{Generalizable Operating Room Re-Identification}

We proceed with our problem statement of how we model short- and long-term personnel tracking through contrastive metric learning in \autoref{subsec:problem_statement}.
Next, we explain our method, which encodes 3D point cloud sequences of OR personnel by capturing the shape and articulated motion while mitigating biases caused by varying textures in the OR environment.
Finally, motivated by the hypothesis that changes in the attire of OR personnel impact the generalizability of commonly studied re-id methods, we outline our framework for re-identifying personnel from RGB- and point cloud-based modalities (\autoref{subsec:re-id_pipeline}), detailing our inference process study in \autoref{subsec:method_inference}.

\subsubsection{Problem Statement}
\label{subsec:problem_statement}

A set of OR recordings $\mathcal{O}$ contains $N$ unique identities.
$\mathcal{O}$ can be partitioned into cropped person sequences $\mathcal{S} = \{\mathcal{S}_i^j \ | \ i\in[N] \;\text{and}\; j \in [m_k] \}$ of length L, where $m_k$ is the number of sequences of person $i \in N$ (see \autoref{fig:main-figure}).
Following a metric learning approach, our goal is to train a deep neural network $F_\Theta$ such that it parameterizes a similarity function $\operatorname{\mathit{sim}}(\cdot, \cdot)$, designed to embed sequences of the same individual onto metrically close points in a low dimensional embedding space $\mathbf{R}^d$.
Specifically, given two different individuals $i$ and $i'$ with respective sequences $j$ and $j'$ we desire a higher similarity between the same individuals:

\[
\begin{aligned}
\operatorname{\mathit{sim}}\left(\mathcal{F}_{\theta}(\mathcal{S}_i^j), \mathcal{F}_{\theta}(\mathcal{S}_{i}^{j'})\right)
&>> \operatorname{\mathit{sim}}\left(\mathcal{F}_{\theta}(\mathcal{S}_i^j), \mathcal{F}_{\theta}(\mathcal{S}_{i'}^{j'})\right). \\
\end{aligned}
\]

Depending on the modality used, a sequence $\mathcal{S}_i^j$ either consists of point clouds $\mathcal{S}_i^j \in \mathbb{R}^{L \times X \times 3}$ with $L$ frames consisting of $X$ total 3D points, or a multi-view RGB image sequence $\mathcal{S}_i^j \in \mathbb{R}^{V \times L \times H \times W \times 3}$ with $L$ frames, and $V$ views with height $H$ and width $W$.

\subsubsection{RGB- and Point Cloud-Based Sequence Encoding}
\label{subsec:re-id_pipeline}

We proceed with how we extract RGB and point cloud input data from our multi-view RGB-D acquisitions (illustrated in \autoref{fig:main-figure}), and then demonstrate our modal-agnostic model architecture and encoding strategy (illustrated in \autoref{fig:main-figure_part_2}). 

\noindent\textbf{Point Cloud Encoding.}
To extract 3D features from a point cloud sequence of each person, we follow the strategy proposed in LidarGait~\cite{lidargait}.
We first use a weakly-supervised approach to segment the point cloud sequences from each acquisition \cite{bastian2023segmentor}.
Next, we center the segmented point clouds (see \autoref{fig:main-figure} in red) of each subject at the world origin and subsequently render depth maps into $V$ virtual camera views ($V=8$) via a perspective projection $\mathcal{P}_{V}: \mathbb{R}^{X \times 3} \mapsto \mathbb{R}^{V \times H \times W \times 1}$, where $X$ is the number of points in the point cloud.
To capture each person from all directions, the $V$ camera views are distributed around each subject in an equidistant circular arrangement (see \autoref{fig:main-figure}).
Finally, each 1-channel depth map is normalized and projected into color space for consistency between the modalities.

We note that LidarGait \cite{lidargait} performs cropping to maximize the content in each virtual view.
This results in a different scaling of each 3D point cloud sequence; we observe that this encoding style adversely affects performance, particularly for generalization to the OR (see \autoref{subsec:results_ablation_over_metric_crop_and_svm}).
Instead, we propose \textit{MetricCrop}, which projects the 3D point clouds into a fixed crop from the floor to a height of 2 meters, as we observe this is the maximum height of subjects in our acquisitions.
This results in a uniformly non-scaled projection, enabling the encoder to encapsulate absolute and not just relative shape differences between individuals (see qualitative examples in our appendix \autoref{fig:metric_crop}).

\noindent\textbf{RGB Encoding.}
To crop the full-sized RGB images to the person, we utilize the person's segmented 3D point cloud and project their 3D bounding box into each of the $V$ RGB views. 
We directly use the 2D RGB crops as input into the 2D encoder, omitting a view if the person is detected as occluded.

\noindent\textbf{Sequence Encoding.}
Inspired by GaitBase \cite{fanOpenGaitRevisitingGait2023}, we encode each frame of a multi-view/temporal input sequence $\mathcal{S}_i^j \in \mathbb{R}^{V \times L \times H \times W \times 3}$ individually with a shared ResNet-9 backbone yielding a total of $L$ features per view $V$ (see \autoref{fig:main-figure_part_2}).
This makes our method agnostic to different camera setups (e.g., a different number of cameras or camera positions), enabling comparisons in various scenarios.
The $V \times L$ features are then aggregated along the temporal dimension through max-pooling.
Each pooled feature is then passed through a horizontal pyramid pooling layer \cite{hpp} to extract information from an individual's body parts separately.
The partial features are then encoded using fully connected layers to generate probe embeddings for each of the $V$ views. 
During training, these probe embeddings are utilized for the triplet loss \cite{leibe_triplet} and, after passing through a batch normalization neck (BNNeck) \cite{bnneck}, are subsequently used for the cross-entropy loss.
We use a variable length $L$ ranging from 10 to 45, as previously shown effective \cite{lidargait}.

\subsubsection{Regularization}
\label{subsubsec:regularization}
For all our experiments, we adopt the \textit{Batchall} variant of the triplet loss \cite{leibe_triplet} with margin $m = 0.2$ to train the encoder, along with a cross-entropy loss weighted with $\lambda = 0.1$
\begin{align*}
\mathcal{L}_{\text{triplet}}(e_0, e_+, e_-) &= \left[ ||e_0 - e_+|| - ||e_0 - e_-|| + m \right]_+ \\
\mathcal{L_\text{full}} &= \mathcal{L}_{\text{triplet}} + \lambda \cdot \mathcal{L}_{\text{ce}},
\end{align*}
\noindent where $e_0$, $e_+$, $e_-$ denote the embeddings under $\mathcal{F}_{\theta}$ of the anchor, positive, and negative sequences, respectively.
As the \textit{Batchall} triplet loss  \cite{leibe_triplet} sums over all combinations of the anchor, positive, and negative pairs, we construct each batch by randomly sampling $K$ instances of $P$ different person IDs, totaling $P \cdot K$ sequences per batch, where $P = 8 \;\text{and} \; K = 8$.

\subsubsection{Inference}
\label{subsec:method_inference}
\noindent\textbf{Multi-View Voting.} 
To obtain a unique embedding, one could combine the view-specific embeddings via a fusion approach \cite{xu2021multi}.
We opt for a simple multi-view voting strategy for both RGB and point cloud-based input, enabling generalization and portability across different environments and datasets.
For one multi-view sequence $\mathcal{S}_i^j$, we obtain a correspondence $\hat{y}_v$ for each view, resulting in a prediction $\hat{Y} = \{\hat{y}_0, ..., \hat{y}_{V-1}\}$.
If more than $\lfloor \frac{V}{2} \rfloor$ correspondences are correct, then $\mathcal{S}_i^j$ is counted as a true positive prediction, and false positive otherwise.
 
\noindent\textbf{Probe-Gallery Inference.} 
\label{subsubsec:probe-gallery-inference}
For inference, we follow the widely adopted probe-gallery protocol \cite{reid_review}.
Given a \textit{gallery} consisting of one sequence per known individual and a \textit{probe} sequence of an unknown individual, we seek to associate the \textit{probe} sequence with a \textit{gallery} sequence via our embedding function $\mathcal{F}_\theta$, uncovering their identity.
Unlike other re-id settings, OR acquisitions are typically long, with many appearances of the same individuals.
We thus propose \textit{SVM-Gallery}, which constructs an OR gallery out of $n$ sequences per individual and trains a \textit{support vector machine} (SVM) \cite{hearst1998support} classifier on the gallery to assign a probe embedding to its best-suited cluster instead of an individual gallery embedding.
We use $n=10$ sequences as we observed performance did not improve significantly for a larger gallery (see appendix for more details).

\subsection{Evaluation Metrics}
\label{subsubsec:evaluation_metrics}

We use various established re-id metrics for our evaluation \cite{ming2022deep}.
\textit{Mean Average Precision} (mAP) calculates the average precision for each probe, which is the average of the precision values at the ranks where correct matches are found. 
The mean of these average precisions is then taken across all queries. 
The rank-3 \textit{cumulative matching characteristic} (CMC@3) estimates the cumulative probability that the correct match is found within the top 3 ranks for each query.
The (rank-1) micro accuracy averages the rank-1 accuracy across all probes, and the (rank-1) macro accuracy first calculates the accuracy per person before averaging these values

In our evaluation, we consider a multi-view setting where each sequence is captured by multiple cameras, with known inter-camera correspondences. This contrasts with single-view approaches that require finding inter-camera correspondences as a separate task \cite{lidargait,huMultiCameraMultiPersonTracking2022a,ni_PartAwareTransformerGeneralizable_2023}.
For each multi-view sequence $\mathcal{S}_i^j$, we obtain $V$ probability vectors $d_v$ (one from each camera) representing the likelihood of a sequence belonging to each identity cluster, as determined by the SVM classifier (see \autoref{subsubsec:probe-gallery-inference}). 
To ensure that our multi-view voting method from \autoref{subsec:method_inference} aligns with the mAP, CMC, and accuracy calculations, we consolidate the $V$ probability vectors into a single vector $m$ using the following reduction strategy.
We begin by initializing $m$ as a zero vector of length equal to the number of identities in the \textit{gallery}. 
The value at the $r$-th position in $m$ is set to 1, where $r$ is the smallest number that satisfies the condition:
\[
\left| \left\{ v \in \{1, 2, \dots, V\} \mid \text{argmax}_i \, d_{v,i} \leq r \right\} \right| > \lfloor \frac{V}{2} \rfloor.
\]
In other words, more than half of the cameras rank the correct identity within the top $r$ positions.

For LidarGait \cite{lidargait}, and PAT \cite{ni2023part}, which utilize a Euclidean nearest-neighbor approach during inference, the $d_v$ vectors are no longer probability vectors but instead represent distance vectors to each instance in the gallery. 
For comparability, we retain the first correct match (i.e., the smallest distance) and then apply the same reduction process as previously described. 
If the $r$-th position exceeds the number of identities, we extend the length of $m$ to accommodate $r$.

We use the same $n=10$ number of sequences per person in the gallery for all methods.

\subsection{Implementation Details.}
We train our models for 40,000 iterations and employ Stochastic Gradient Descent (SGD) as an optimizer with a weight decay of 0.0005 and a learning rate of 0.1, which is decreased by a factor of 10 every 10,000 iterations.
All networks are trained on a single Nvidia RTX 2080 Ti with 11GB VRAM, using PyTorch 1.13.0.

For each train-test setup, we tune hyperparameters on a random shuffle of the dataset and select the best-performing data augmentation strategy accordingly (see appendix for more details).
For RGB-based methods, we use horizontal flipping and random cropping to increase sample diversity and local grayscale transformation \cite{gong2021eliminate}, which randomly converts image patches to grayscale, encouraging the networks to learn more robust features beyond color information.
For point cloud-based methods, we use random erasing \cite{zhong2020random} and Gaussian noise to improve the model's robustness against partial point clouds.
For PAT \cite{ni2023part} we use the code published by the authors and follow their code base to train the models.

\subsection{3D OR Activity Imprints}

We propose 3D activity imprints for visualizating macroscopic OR personnel behavior.
Our visualization pipeline proceeds as follows: we first segment a single, complete 3D point cloud of the OR using weakly-supervised 3D point cloud segmentation \cite{bastian2023segmentor}.
For the base OR layout visualization, we orthographically project all 3D points classified as ``background'' into a bird's-eye view and overlay the person's activity patterns.
To visualize these activity patterns, we aggregate each person's re-identified 3D point cloud sequence and generate an activity imprint where the color reflects the spatial occupancy frequency throughout the procedure. 
By additionally incorporating the segmented OR objects (e.g., the patient table, or other equipment) into the bird's-eye projection, we can also automatically adapt to different OR layouts. 

\section{Experiments and Results}

\subsection{Datasets}
\label{subsubsec:datasets}

For evaluation, we use the publicly available 4D-OR dataset \cite{ozsoy20224d}, our internal OR dataset \ordataset, as well as the general streetwear re-identification dataset SUSTech1k \cite{lidargait}.
What follows is an overview of the three datasets used in \autoref{tbl:datasets_overview}, including the number of individuals present.

\begin{figure*}[hbt]
    \centering
    \includegraphics[width=\textwidth]{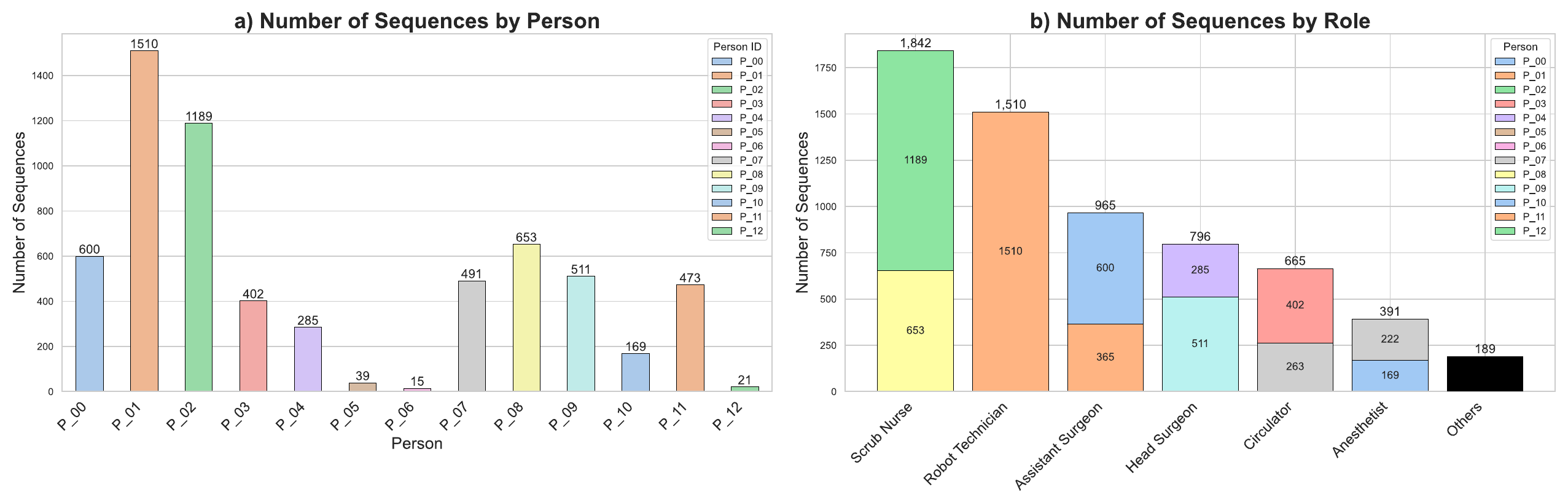}
    \caption{
    A statistical overview depicting the distribution of roles and persons in \ordataset.
    a) shows the distribution of sequences per person.
    b) shows the distribution of sequences and individuals per role. ``Others'' denotes individuals in recordings that do not correspond to any of the six roles.
    }
    \label{fig:data_plots}
\end{figure*}

\textbf{4D-OR} \cite{ozsoy20224d} consists of 10 simulated total knee replacement surgeries, recorded with six RGB-D sensors.
Five roles are simulated by five distinct individuals: head surgeon, assistant surgeon, circulator, anesthetist, and patient.
The acquisitions exhibit relatively little variance in appearance, and each individual wears distinct streetwear under their OR gowns, yielding clearly identifiable features (see \autoref{fig:saliency_maps}).
For our experiments, we use the 15 FPS version of 4D-OR to obtain higher-resolution motion sequences.

\textbf{SUSTech1K} \cite{lidargait} is a large-scale Lidar-based gait recognition dataset.
It contains 25,279 walking sequences of 1,050 individuals.
Each walking sequence contains 10-45 frames at 10 FPS.
We use the dataset's available labels, RGB sequences, and rendered point cloud sequences.

\textbf{\ordataset} consists of 11 simulated robotic (left) knee surgeries, recorded over six months, involving professional OR personnel, all of whom were dressed in role-specific OR attire. 
The dataset includes 13 unique individuals, with each surgery featuring six distinct roles: head surgeon, assistant surgeon, scrub nurse, circulator, robotic technician, and anesthetist. 
As depicted in \autoref{fig:data_plots} a), the amount of data per individual varies, reflecting the differing durations of the distinct roles in the OR, as shown in \autoref{fig:data_plots} b).
The \ordataset was internally acquired using five co-registered, ceiling-mounted RGB-D Azure Kinect cameras (see the camera setup in \autoref{fig:main-figure}), capturing at 15 FPS.
Given that some surgeries extend up to 2 hours, we annotate approximately 70\% of each procedure, generating 6,358 sequences.
Each frame contains both a 3D point cloud and multi-view RGB images. The frame counts across all sequences are uniformly distributed between 10 and 45 frames.
While datasets like 4D-OR and SUSTech1K provide ground-truth sequence labels, we employ a weakly supervised approach \cite{bastian2023segmentor} to derive similar sequence labels.
Further qualitative examples of RGB and rendered 3D point clouds of \ordataset are depicted in the appendix.

\begin{table}
\centering
\caption{Overview of datasets used in the study. \textbf{IDs} denotes the number of unique identities present, and the number of annotated \textbf{Sequences} in each dataset.}
\resizebox{\columnwidth}{!}{%
\begin{tabular}{lccccc}
\toprule
\textbf{Dataset} & \textbf{Domain} & \textbf{\# IDs} & \textbf{\# Sequences} & \textbf{\# Cameras} & \textbf{FPS} \\
\midrule
SUSTech1K \cite{lidargait} & General Outdoor & 1,050 & 25,279 & 12 & 10 \\ %
4D-OR \cite{ozsoy20224d}  & Surgical OR & 5  & 3,734  & 6 & 15 \\
\ordataset  & Surgical OR & 13  & 6,358  & 5 & 15 \\
\bottomrule
\end{tabular}%
}
\label{tbl:datasets_overview}
\end{table}

\subsection{Experiment 1: RGB-Based Texture-bias in Re-ID}
\label{subsec:results_deficits_of_rgb_based_person_re-identification}

\subsubsection{Setup}
To systematically analyze the visual features CNNs exploit for RGB-based person re-identification in the OR, we turn to saliency maps, which use gradients and feature activations to depict important regions for model decision-making 
(see \autoref{fig:saliency_maps}).
We train a ResNet-50 CNN on a collection of 224x224 cropped RGB images with a cross-entropy loss, applying GradCAM \cite{selvaraju2017grad} to visualize the class activation maps in the final convolutional layer.

\subsubsection{Findings}
The saliency maps in \autoref{fig:saliency_maps} illustrate how the CNN processes RGB images differently for 4D-OR and \ordataset.
In 4D-OR, GradCAM \cite{selvaraju2017grad} suggests the CNNs attend to specific anatomical regions, particularly the head and feet areas of the individuals across all images.
In contrast, such distinct patterns are not visible for \ordataset: the saliency maps here appear more diffused, lacking distinct regions of focus as observed in 4D-OR.

\subsection{Experiment 2: Human-Pose-Based Tracking}
\label{subsec:results_human_pose_based_tracking}
\subsubsection{Setup}

We evaluate our method against existing 3D human-pose-based tracking approaches on a complete surgery from our dataset. 
For comparison, we implement two baseline tracking methods: a naive tracking approach that uses linear assignment between frames and the KSP tracker \cite{KSP_tracker}, which leverages globally optimal paths.
The naive tracking method maintains a memory bank of each person's last known pose to handle temporary track losses from either missed detections or when people exit the OR. 
When tracks are lost, it attempts to re-associate them by solving a linear assignment problem between new pose estimates and the stored memory bank poses using minimum cost bipartite matching \cite{crouse_on_implementing_2d_rectangular} as in \cite{ozsoy20224d}. 
The KSP tracker takes a different approach, generating 2D occupancy maps from the 3D pose estimates and using OR doorways as access points to track paths.
Both baseline methods require knowing the total number of people present and their initial appearances in the surgery. 
Since our dataset only includes annotations for $\sim70\%$ of each procedure (see \autoref{subsubsec:datasets}), we additionally annotate all remaining appearances through this specific surgery.

To gauge the usefulness of augmenting human pose trackers with re-identification features, we construct a re-identification-based tracker by first aggregating each person's 3D point clouds into short sequences (10-45 frames) using linear assignment, then continuously associate these sequences with identities from a pre-constructed gallery.
The 3D human pose estimates used by all methods are obtained from a backbone network \cite{Srivastav_2024_CVPR} trained in a self-supervised manner on the remaining surgeries in our dataset, while our re-identification model is trained on two separate surgeries.

\begin{table}
    \normalsize
    \centering
    \caption{Rank-1 micro and macro accuracy (in percentage) when tracking individuals over an entire surgery of \ordataset based on associating 3D human poses between frames (Naive Tracking, KSP Tracker) compared to re-identification based tracking [Ours (PC, $n$)], where $n$ denotes the number of sequences per person in the gallery.
    }
    \resizebox{.8\columnwidth}{!}{%
    \label{tab:simple_tracking_small}
    \begin{tabular}{l|cc}
        \toprule
        \textbf{Method} & Avg. Micro & Avg. Macro \\
        \hline
        Naive Tracking \cite{ozsoy20224d} & 15.79 & 33.01 \\
        KSP Tracker \cite{KSP_tracker} & 19.83 & 16.49 \\ 
        \hline
        Ours (PC, $1$) & 55.84 & 58.51 \\
        Ours (PC, $5$) & 73.30 & 71.80 \\
        Ours (PC, $10$) & 76.54 & 75.28 \\
        \bottomrule
    \end{tabular}
    }
\end{table}

\subsubsection{Findings}
\autoref{fig:joints_3d_figure} depicts a qualitative example sequence analyzing the naive 3D human-pose-based tracking approach.
The pose-estimation accuracy varies between individuals, as \textit{P\_02} and \textit{P\_17} show fewer missed detections (indicated by fewer gray bars), while more frequent detection failures occur for individuals \textit{P\_00} and \textit{P\_15}.
Moreover, an incorrect identity assignment between \textit{P\_00} and \textit{P\_15} midway through the sequence propagates into subsequent frames, resulting in persistent tracking errors (shown as red bars) for the remainder of the sequence.

In \autoref{tab:simple_tracking_small}, we report the quantitative results in average micro and macro accuracy of the naive 3D human pose-based-tracker, KSP tracker \cite{KSP_tracker}, and our person re-ID-based tracker.
The naive 3D human pose-tracker and KSP-tracker achieve an average micro accuracy of 16\% and 20\%, respectively.
In comparison, our re-ID-based tracker achieves a micro accuracy of 56\% with a single annotated sequence in the gallery, which increases to 73\% with 5 annotated sequences and 77\% with 10 annotated sequences per individual.
The macro accuracies follow a similar trend as the micro accuracies; however, for the naive tracking method, the macro accuracy improves from 15.79\% to 33.01\%.

\begin{table*}[h!t]
\centering
\setlength{\tabcolsep}{8pt}  %
\renewcommand{\arraystretch}{1.1} %
\caption{
Comparison of inter- and intra-dataset performance between LidarGait \cite{lidargait}, PAT \cite{ni2023part}, and our method. Methods are color-coded based on the input modality: \colorbox{rgbexp!15}{RGB} for RGB input and \colorbox{depthexp!15}{point cloud (PC)} for point cloud input. We evaluate performance across three datasets: the two OR datasets, \ordataset and 4D-OR \cite{ozsoy20224d}, and the general computer vision dataset, SUSTech1K \cite{lidargait}. 
The metrics include mean average precision (mAP), rank-3 cumulative matching characteristics (CMC@3), rank-1 micro accuracy [Acc. (Micro)], and rank-1 macro accuracy [Acc. (Macro)].
Each metric reports the average and standard error across the four-fold cross-validation in percentage.
The best value for each metric is highlighted in \textbf{bold}. 
Statistical significance with respect to the best performing method is shown for each other method (``***": $p < 0.001$, ``**": $p < 0.01$, ``*": $p < 0.05$, ``ns'': not significant).}
\label{tbl:acc_performance}
\resizebox{\linewidth}{!}{
\begin{tabular}
{
  @{}
  l|
  l|
  c|
  c|
  >{\raggedright\arraybackslash}p{3.5cm}|
  >{\raggedright\arraybackslash}p{3.5cm}|
  >{\raggedright\arraybackslash}p{3.5cm}|
  >{\raggedright\arraybackslash}p{3.5cm}|
  @{}
}
\toprule
& \multicolumn{1}{l|}{\textbf{Method}} &
  \multicolumn{1}{l|}{\textbf{Train}} &
  \multicolumn{1}{l|}{\textbf{Test}} &
  \textbf{mAP} &
  \textbf{CMC@3} &
  \textbf{Acc. (Micro)} &
  \textbf{Acc. (Macro)} \\
\midrule
\parbox[t]{2mm}{\multirow{8}{*}{\rotatebox[origin=c]{90}{\large \textit{Intra Dataset}}}}
& \multicolumn{1}{l|}{\rgbcell PAT} &
  \multicolumn{2}{c|}{\multirow{4}{*}{4D-OR}} & \rgbcell
    $97.43 \pm 0.37^{\text{ns}}$ & \rgbcell $97.61 \pm 0.32^{*}$ & \rgbcell $96.00 \pm 0.61^{\text{ns}}$ & \rgbcell $96.11 \pm 0.74^{\text{ns}}$ \\
& \multicolumn{1}{l|}{\rgbcell Ours (RGB)} &
  \multicolumn{2}{c|}{} & \rgbcell
    $\mathbf{98.57 \pm 0.30}$ & \rgbcell $\mathbf{98.97 \pm 0.22}$ & \rgbcell $\mathbf{97.24 \pm 0.57}$ & \rgbcell $\mathbf{96.91 \pm 0.72}$ \\
& \multicolumn{1}{l|}{\depthcell LidarGait} &
  \multicolumn{2}{c|}{} & \depthcell 
  $94.20 \pm 0.77^{*}$ & \depthcell $94.02 \pm 0.77^{**}$ & \depthcell $92.34 \pm 1.14^{*}$ & \depthcell $92.17 \pm 1.12^{*}$ \\
& \multicolumn{1}{l|}{\depthcell Ours (PC)} &
  \multicolumn{2}{c|}{} & \depthcell 
  $97.84 \pm 0.16^{\text{ns}}$ & \depthcell $98.29 \pm 0.16^{\text{ns}}$ & \depthcell $95.87 \pm 0.43^{\text{ns}}$ & \depthcell $95.95 \pm 0.45^{\text{ns}}$ \\
\cmidrule(l){2-8}
& \multicolumn{1}{l|}{\rgbcell PAT} &
  \multicolumn{2}{c|}{\multirow{4}{*}{\ordataset}} &
     \rgbcell $82.56 \pm 0.95^{*}$ & \rgbcell $82.55 \pm 0.95^{**}$ & \rgbcell $74.35 \pm 1.51^{*}$ & \rgbcell $77.68 \pm 1.34^{*}$ \\
& \multicolumn{1}{l|}{\rgbcell Ours (RGB)} &
  \multicolumn{2}{c|}{} & \rgbcell
    $81.72 \pm 2.23^{*}$ & \rgbcell $84.02 \pm 2.34^{*}$ & \rgbcell $69.50 \pm 3.11^{*}$ & \rgbcell $73.23 \pm 3.13^{*}$ \\
& \multicolumn{1}{l|}{\depthcell LidarGait} &
  \multicolumn{2}{c|}{} & \depthcell 
  $88.00 \pm 1.99^{\text{ns}}$ & \depthcell $87.57 \pm 2.07^{\text{ns}}$ & \depthcell $84.31 \pm 2.61^{\text{ns}}$ & \depthcell $83.62 \pm 2.10^{\text{ns}}$ \\
& \multicolumn{1}{l|}{\depthcell Ours (PC)} &
  \multicolumn{2}{c|}{} & \depthcell
    $\mathbf{91.58 \pm 1.33}$ & \depthcell $\mathbf{92.45 \pm 1.23}$ & \depthcell $\mathbf{86.19 \pm 1.94}$ & \depthcell $\mathbf{85.74 \pm 1.69}$ \\
\midrule
\midrule
\parbox[t]{2mm}{\multirow{16}{*}{\rotatebox[origin=c]{90}{\large \textit{Inter Dataset}}}}
& \multicolumn{1}{l|}{\rgbcell PAT} &
  \multicolumn{1}{c|}{\multirow{4}{*}{\ordataset}} &
  \multicolumn{1}{c|}{\multirow{4}{*}{4D-OR}} & \rgbcell
    $79.81 \pm 1.48^{**}$ & \rgbcell $79.79 \pm 1.66^{**}$ & \rgbcell $69.21 \pm 1.94^{**}$ & \rgbcell $71.15 \pm 1.41^{**}$ \\
& \multicolumn{1}{l|}{\rgbcell Ours (RGB)} &
  \multicolumn{1}{c|}{} &
  \multicolumn{1}{c|}{} & \rgbcell
    $73.02 \pm 1.47^{***}$ & \rgbcell $76.13 \pm 1.56^{**}$ & \rgbcell $53.10 \pm 2.33^{***}$ & \rgbcell $54.69 \pm 2.69^{**}$ \\
& \multicolumn{1}{l|}{\depthcell LidarGait} &
  \multicolumn{1}{c|}{} &
  \multicolumn{1}{c|}{} & \depthcell 
  $80.14 \pm 1.24^{**}$ & \depthcell $79.32 \pm 1.29^{**}$ & \depthcell $72.82 \pm 1.68^{**}$ & \depthcell $74.07 \pm 1.81^{**}$ \\
& \multicolumn{1}{l|}{\depthcell Ours (PC)} &
  \multicolumn{1}{c|}{} &
  \multicolumn{1}{c|}{} & \depthcell 
    $\mathbf{94.09 \pm 0.23}$ & \depthcell $\mathbf{95.28 \pm 0.22}$ & \depthcell $\mathbf{89.25 \pm 0.31}$ & \depthcell $\mathbf{89.43 \pm 0.33}$ \\ 
\cmidrule(l){2-8}
& \multicolumn{1}{l|}{\rgbcell PAT} &
  \multicolumn{1}{c|}{\multirow{4}{*}{4D-OR}} &
  \multicolumn{1}{c|}{\multirow{4}{*}{\ordataset}} & \rgbcell
    $63.62 \pm 0.47^{***}$ & \rgbcell $62.19 \pm 0.56^{***}$ & \rgbcell $48.66 \pm 0.64^{**}$ & \rgbcell $54.72 \pm 1.53^{**}$ \\
& \multicolumn{1}{l|}{\rgbcell Ours (RGB)} &
  \multicolumn{1}{c|}{} &
  \multicolumn{1}{c|}{} & \rgbcell
    $60.48 \pm 0.76^{***}$ & \rgbcell $59.42 \pm 0.75^{***}$ & \rgbcell $40.14 \pm 1.11^{***}$ & \rgbcell $46.97 \pm 2.53^{**}$ \\
& \multicolumn{1}{l|}{\depthcell LidarGait} &
  \multicolumn{1}{c|}{} &
  \multicolumn{1}{c|}{} & \depthcell 
  $62.95 \pm 1.29^{**}$ & \depthcell $61.63 \pm 1.41^{**}$ & \depthcell $50.01 \pm 1.80^{**}$ & \depthcell $54.72 \pm 1.14^{**}$ \\
& \multicolumn{1}{l|}{\depthcell Ours (PC)} &
  \multicolumn{1}{c|}{} &
  \multicolumn{1}{c|}{} &
    \depthcell $\mathbf{77.65 \pm 0.45}$ & \depthcell $\mathbf{79.08 \pm 0.26}$ & \depthcell $\mathbf{63.13 \pm 1.25}$ & \depthcell $\mathbf{65.93 \pm 0.64}$ \\
\cmidrule(l){2-8}
& \multicolumn{1}{l|}{\rgbcell PAT} &
  \multicolumn{1}{c|}{\multirow{4}{*}{SUSTech1K}} &
  \multicolumn{1}{c|}{\multirow{4}{*}{4D-OR}} & \rgbcell
    $83.12 \pm 0.77^{***}$ & \rgbcell $83.56 \pm 0.79^{***}$ & \rgbcell $73.61 \pm 1.15^{***}$ & \rgbcell $74.78 \pm 0.79^{***}$ \\
& \multicolumn{1}{l|}{\rgbcell Ours (RGB)} &
  \multicolumn{1}{c|}{} &
  \multicolumn{1}{c|}{} & \rgbcell
    $79.67 \pm 0.77^{***}$ & \rgbcell $82.55 \pm 0.79^{***}$ & \rgbcell $64.36 \pm 1.22^{***}$ & \rgbcell $65.14 \pm 1.13^{***}$ \\
& \multicolumn{1}{l|}{\depthcell LidarGait} &
  \multicolumn{1}{c|}{} &
  \multicolumn{1}{c|}{} & \depthcell 
  $75.55 \pm 1.13^{***}$ & \depthcell $74.77 \pm 1.23^{***}$ & \depthcell $65.77 \pm 1.48^{***}$ & \depthcell $67.64 \pm 1.58^{***}$ \\
& \multicolumn{1}{l|}{\depthcell Ours (PC)} &
  \multicolumn{1}{c|}{} &
  \multicolumn{1}{c|}{} &
    \depthcell $\mathbf{94.13 \pm 0.27}$ & \depthcell $\mathbf{95.07 \pm 0.24}$ & \depthcell $\mathbf{89.87 \pm 0.29}$ & \depthcell $\mathbf{90.06 \pm 0.26}$ \\
\cmidrule(l){2-8}
& \multicolumn{1}{l|}{\rgbcell PAT} &
  \multicolumn{1}{c|}{\multirow{4}{*}{SUSTech1K}} &
  \multicolumn{1}{c|}{\multirow{4}{*}{\ordataset}} & \rgbcell
    $74.27 \pm 0.39^{**}$ & \rgbcell $73.69 \pm 0.39^{***}$ & \rgbcell $62.87 \pm 0.60^{**}$ & \rgbcell $68.18 \pm 0.82^{**}$ \\
& \multicolumn{1}{l|}{\rgbcell Ours (RGB)} &
  \multicolumn{1}{c|}{} &
  \multicolumn{1}{c|}{} & \rgbcell
    $72.76 \pm 0.49^{**}$ & \rgbcell $72.94 \pm 0.62^{**}$ & \rgbcell $57.66 \pm 0.54^{**}$ & \rgbcell $62.82 \pm 1.86^{**}$ \\
& \multicolumn{1}{l|}{\depthcell LidarGait} &
  \multicolumn{1}{c|}{} &
  \multicolumn{1}{c|}{} & \depthcell 
  $70.30 \pm 0.59^{***}$ & \depthcell $69.40 \pm 0.69^{***}$ & \depthcell $59.52 \pm 0.81^{**}$ & \depthcell $63.01 \pm 0.80^{**}$ \\
& \multicolumn{1}{l|}{\depthcell Ours (PC)} &
  \multicolumn{1}{c|}{} &
  \multicolumn{1}{c|}{} &
    \depthcell $\mathbf{84.45 \pm 0.88}$ & \depthcell $\mathbf{85.01 \pm 0.80}$ & \depthcell $\mathbf{75.27 \pm 1.47}$ & \depthcell $\mathbf{76.63 \pm 0.76}$ \\
\bottomrule
\end{tabular}
}
\end{table*}

\subsection{Experiment 3: Intra and Inter-OR Re-identification}
\label{subsec:results_experiment_generalization}

\subsubsection{Setup}
Method names are colored according to their input modality (\colorbox{rgbexp!15}{RGB} or \colorbox{depthexp!15}{point-cloud}) in this subsection to aid interpretation.

We evaluate the re-identification performance between \colorbox{depthexp!15}{LidarGait} \cite{lidargait}, the transformer-based domain generalizable person re-identification model \colorbox{rgbexp!15}{PAT} \cite{ni2023part}, our method using RGB as input \colorbox{rgbexp!15}{Ours (RGB)}, and our method using 3D point cloud sequences as input \colorbox{depthexp!15}{Ours (PC)} on all three datasets introduced in \autoref{subsubsec:datasets}. 
For \colorbox{rgbexp!15}{PAT} \cite{ni2023part}, we sample the mid-point frame of each sequence, as it only takes a single frame as input.

We perform four-fold cross-validation, partitioning folds by surgery. 
Note that while we have annotated 6,358 and 3,734 sequences in \ordataset and 4D-OR, respectively, our experiments use 5,504 and 3,719 sequences after removing cases where individuals were heavily occluded in RGB images to ensure consistency between RGB and depth modalities (see appendix for the exact number of instances per fold).
The average and standard errors are reported across the four splits.
We additionally conduct paired t-tests and report statistical significance between the best method and each other method.

To assess the generalization capability of the methods across different OR environments, we conduct both intra-domain and inter-domain experiments.
For SUSTech1K \cite{lidargait}, we use the train split provided by the authors, evaluating a different fold of an OR dataset for each split.
The model generalization performance is assessed as is, we do not assess domain adaptation techniques.

\begin{figure*}[ht]
    \centering
    \includegraphics[width=\textwidth]{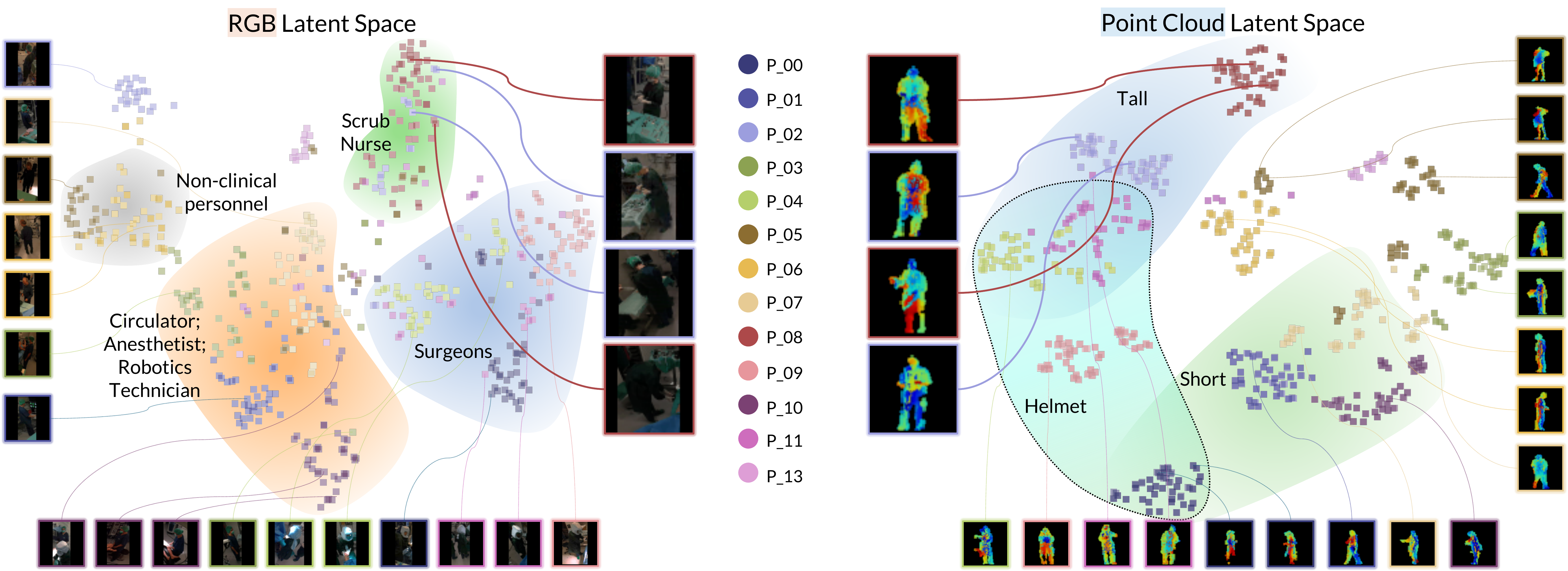}
    \caption{
    Visualization of the \colorbox{rgbexp!15}{RGB} and \colorbox{depthexp!15}{point cloud} latent spaces on \ordataset, visualized by a t-SNE projection.
    Regions are manually highlighted based on the common attributes within clusters in the latent spaces.
    }
    \label{fig:qual-results}
    \vspace{-0.3cm}
\end{figure*}

\subsubsection{Findings}
\autoref{tbl:acc_performance} presents a detailed performance overview across methods, metrics, datasets, and modalities.

\textbf{Intra-Domain.} 
Intra-dataset comparisons on 4D-OR \cite{ozsoy20224d} depict the performance of \colorbox{rgbexp!15}{PAT} \cite{ni2023part} [96.11\% Acc. (Macro)], \colorbox{rgbexp!15}{Ours (RGB)} [96.91\% Acc. (Macro)], \colorbox{depthexp!15}{LidarGait} \cite{lidargait} [92.17\% Acc. (Macro)], and \colorbox{depthexp!15}{Ours (PC)} [95.95\% Acc. (Macro)] as similar, with both RGB and point cloud modalities achieving high scores across all metrics.
However, on \ordataset, the point cloud-based methods performance exceeds those of the RGB-based methods by a small yet statistically significant margin. 
For instance, \colorbox{depthexp!15}{Ours (PC)} outperforms \colorbox{rgbexp!15}{Ours (RGB)} by 12\%, and \colorbox{depthexp!15}{LidarGait} outperforms \colorbox{rgbexp!15}{Ours (RGB)} by 10\% in Acc. (Macro).
Although the point cloud-based methods [\colorbox{depthexp!15}{LidarGait} and \colorbox{depthexp!15}{Ours (PC)}] outperform \colorbox{rgbexp!15}{PAT} on \ordataset, \colorbox{rgbexp!15}{PAT} notably achieves a 4\% higher Acc. (Macro) than \colorbox{rgbexp!15}{Ours (RGB)}.

\textbf{Generalization.} 
Our cross-dataset performance analysis, evaluated across \ordataset, 4D-OR \cite{ozsoy20224d}, and SUSTech1K  \cite{lidargait}, reveals a disparity in the generalization capabilities between \colorbox{depthexp!15}{Ours (PC)} and the other methods.
Notably, \colorbox{depthexp!15}{Ours (PC)} demonstrates consistently high performance across different datasets, significantly outperforming \colorbox{rgbexp!15}{Ours (RGB)}, \colorbox{rgbexp!15}{PAT} \cite{ni2023part} and \colorbox{depthexp!15}{LidarGait} \cite{lidargait} ($p < 0.01$). 
For instance, when cross-trained and tested on either \ordataset and 4D-OR, \colorbox{depthexp!15}{Ours (PC)} achieves superior performance, with all metrics consistently higher than \colorbox{depthexp!15}{LidarGait}, \colorbox{rgbexp!15}{PAT} \cite{ni2023part}, and \colorbox{rgbexp!15}{Ours (RGB)} (11\%+ in macro accuracy). 

In contrast, when tested on out-of-domain datasets, \colorbox{rgbexp!15}{Ours (RGB)} experiences significant performance drops, indicating potential overfitting to dataset-specific characteristics. 
This trend is particularly evident when \colorbox{rgbexp!15}{Ours (RGB)} is trained on the small-scaled 4D-OR dataset and tested on \ordataset, showing a marked decline in performance, with an Acc. (Macro) of 46.97\% compared to \colorbox{depthexp!15}{Ours (PC)} at 65.93\%.

\colorbox{rgbexp!15}{PAT} \cite{ni2023part} demonstrates strong cross-dataset performance in specific scenarios, often outperforming \colorbox{depthexp!15}{LidarGait} \cite{lidargait} but still exceeded by the results of \colorbox{depthexp!15}{Ours (PC)} ($p < 0.01$). 
For instance, when trained on the large-scale SUSTech1K dataset \cite{lidargait} and tested on \ordataset, \colorbox{rgbexp!15}{PAT} achieves an Acc. (Macro) of 68.18\%, above \colorbox{depthexp!15}{LidarGait's} performance of 63.01\% yet significantly below \colorbox{depthexp!15}{Ours (PC)'s} performance of 76.63\%. 
However, in some cases, such as when trained on \ordataset and tested on 4D-OR, \colorbox{rgbexp!15}{PAT's} performance [Acc. (Macro) of 71.15\%] falls even further behind \colorbox{depthexp!15}{Ours (PC)} [Acc. (Macro) of 89.43\%].

\textbf{Latent Embedding Visualization.}
To further validate that the embeddings of \colorbox{rgbexp!15}{Ours (RGB)} and \colorbox{depthexp!15}{Ours (PC)} focus on meaningful features for person re-identification, we also analyze the latent space (see \autoref{fig:qual-results}) using dimension reduction with t-SNE \cite{van2008visualizing}.
While the RGB latent space is more scattered, the point cloud latent space appears to be better separated.
For instance, \textit{P\_02} and \textit{P\_08} appear to be entangled in the RGB latent space, whereas \textit{P\_02} and \textit{P\_08} are demarcated more clearly as two distinct clusters in the point cloud latent space.

\subsection{Experiment 4: Downstream Analysis of OR Workflows through 3D Activity Imprints}

\begin{figure*}[htbp]
    \centering
    \includegraphics[width=.95\textwidth]{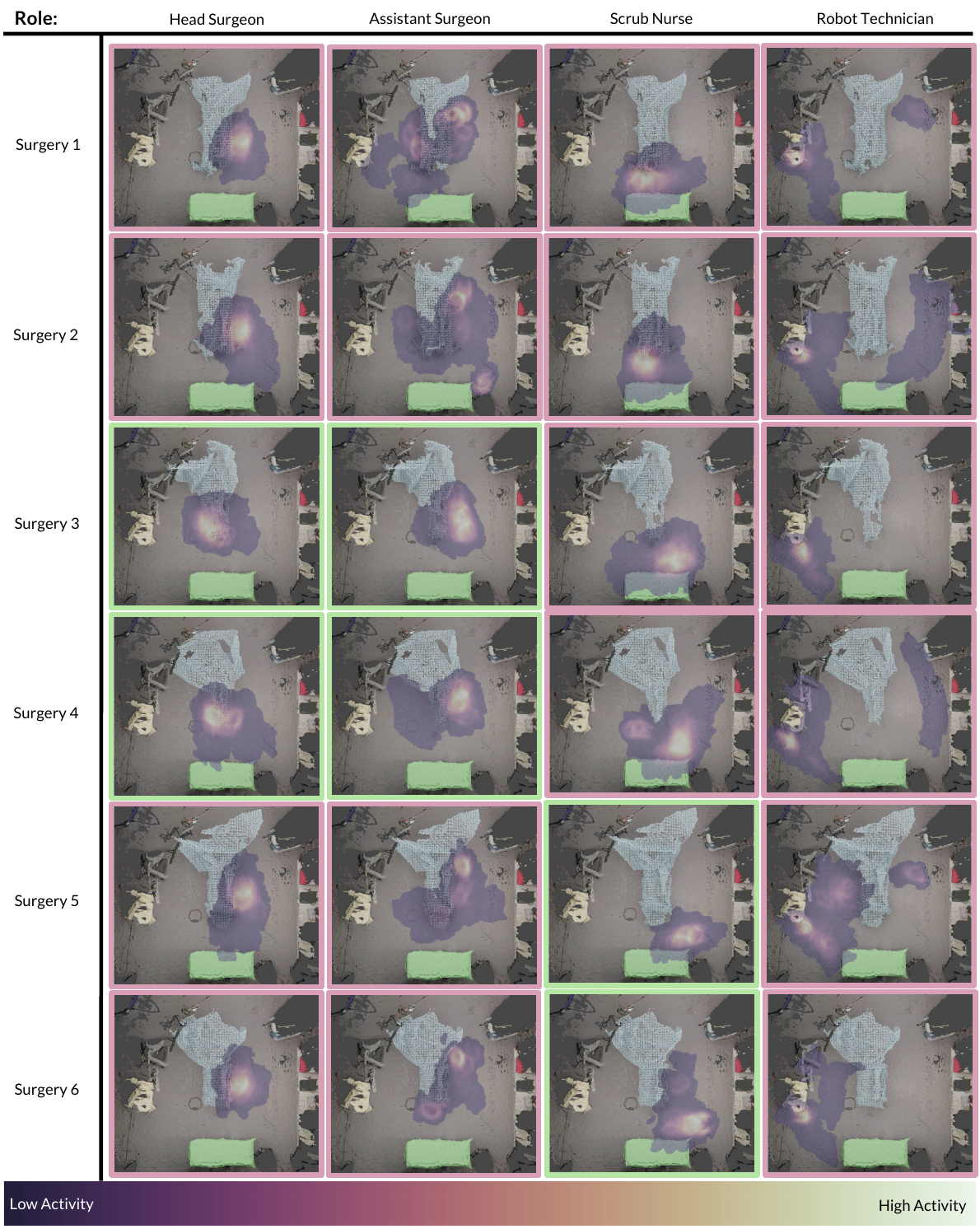}
    \caption{3D OR activity imprints generated via automated personnel tracking during the intra-operative phase. Each role is represented by a maximum of two individuals, distinguished by borders of different colors (\colorbox{heatmap_green!70}{green} and \colorbox{heatmap_pink!70}{pink}). 
    The patient table is highlighted in \colorbox{patient_table!70}{blue}, the tool table in \colorbox{tool_table!70}{green}, and the robot maintenance station in \colorbox{mps_station!70}{gold}.
    The data is generated from \ordataset.}
    \label{fig:heatmaps}
\end{figure*}

\subsubsection{Setup}
We use our best-performing method to automatically analyze the activity patterns of various roles during a procedure to create 3D activity imprints on several surgeries, roles, and staff members of \ordataset.
The location of each staff member is visualized using a continuous colormap depicting low to high activity.
When plotting multiple staff members on a scene, we use distinct colormaps for each individual.

\subsubsection{Findings} 
\autoref{fig:heatmaps} depicts our 3D activity imprints of the intra-operative phase for six different surgeries, highlighting the movements of four distinct roles performed by seven staff members.
Our 3D activity imprints depict the OR, including the patient table highlighted in blue, the tool table highlighted in green, and the robot maintenance station highlighted in gold.

The 3D activity imprints indicate that the head surgeons (1st column) remain relatively stationary on either side of the operating table while the assistant surgeons (2nd column) move between both sides. 
The scrub nurses (3rd column) primarily stay between the tool table and the head surgeon, indicating a more stationary role.
In contrast, the robotics technician (4th column) demonstrates more extensive movements across the room.

A deeper examination of the head surgeons' activity pattern shows that the \colorbox{heatmap_green!70}{green head surgeon} (denoted by a green border in \autoref{fig:heatmaps}) consistently stands on the patient's left side during surgeries 3 and 4, whereas the \colorbox{heatmap_pink!70}{pink head surgeon} predominantly operates from the right side of the patient table in surgeries 1, 2, 5, and 6.

While the scrub nurses stay relatively stationary between the patient table and the tool table, interestingly, the \colorbox{heatmap_pink!70}{pink scrub nurse} occupies different locations in surgery 1 and 2 compared to surgery 3 and 4.

\begin{figure*}[ht]
    \centering
    \includegraphics[width=\textwidth]{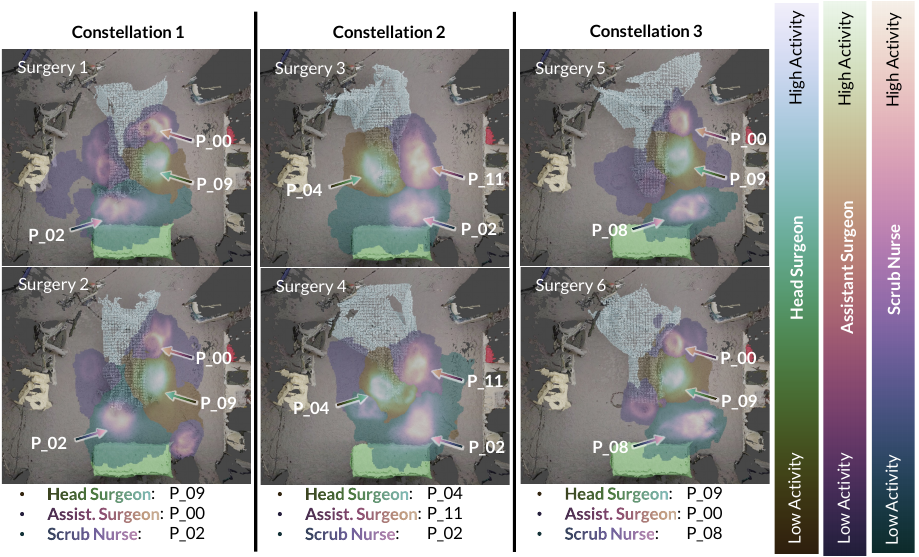}
    \caption{Different personnel constellations and their respective 3D activity imprints. 
    Our proposed re-ID-based tracking approach yields insight into the coordination of surgical teams, providing insight into group workflow patterns and usage of the OR for a given surgery.
    The patient table is visualized in \colorbox{patient_table!70}{blue}; the tool-table in \colorbox{tool_table!70}{green}; the robot maintenance station in \colorbox{mps_station!70}{gold}.
    The data is generated from \ordataset.
    }
    \label{fig:const_heatmaps}
\end{figure*}

\autoref{fig:const_heatmaps} depicts an alternative form of 3D activity imprint visualization, where the activity patterns of multiple individuals are aggregated to portray an overview of the surgical workflow. 
Each activity imprint includes the activity of the head surgeon, assistant surgeon, and scrub nurse, with distinct colormaps used to differentiate the roles. 
This combined representation provides a compact and effective means to compare different surgical configurations and their preferred workflows.

Constellation 2 (2nd column) has high activity clusters that are localized for the head surgeon (P\_04), the assistant surgeon (P\_01), and the scrub nurse (P\_02). 
In constellation 1, the activities are more spread out, while in constellation 3, the left side of the patient table is used less. 
Furthermore, in constellation 3, the assistant surgeon has two high-activity centers with the head surgeon positioned between them.

\subsection{Experiment 5: Role Prediction via Person Re-Id}

\begin{table*}[ht]
\centering
\setlength{\tabcolsep}{3pt} %
\caption{Role prediction on 4D-OR \cite{ozsoy20224d}. We compare our re-identification approach against scene graph-based role prediction \cite{ozsoy20224d}. Numbers are reported in percentage of accuracy. ``-'' indicates the role is not present in that phase.}
\label{tbl:abl_role_prediction}
\begin{tabular}{c|ccc|ccc|ccc|ccc}
\toprule
& \multicolumn{3}{c|}{Head Surgeon} & \multicolumn{3}{c|}{Assistant Surgeon} & \multicolumn{3}{c|}{Circulating Nurse} & \multicolumn{3}{c}{Anesthetist} \\
\cmidrule{2-13}
Method & Pre & Intra & Post & Pre & Intra & Post & Pre & Intra & Post & Pre & Intra & Post \\
\midrule
4D-OR \cite{ozsoy20224d}    & 92.28 & 100.0 & - & 90.79 & 51.96 & - & 60.67 & 47.99 & 86.19 & 63.19 & 59.63 & 67.65 \\
Ours (Depth) & 100.0 & 100.0 & - & 100.0 & 99.50 & - & 99.56 & 100.0 & 100.0 & 85.64 & 100.0 & 93.11 \\
\bottomrule
\end{tabular}
\end{table*}

\subsubsection{Setup}
In 4D-OR \cite{ozsoy20224d}, the authors propose scene graphs to predict surgical roles.
However, scene graphs rely heavily on human-object interactions and fail when these are missing or incorrectly detected.
To demonstrate these shortcomings, we stratify role predictions across the surgical phases provided in the dataset.
Following the methodology of \cite{ozsoy20224d}, we annotate ground truth role tracks for each individual covering the given surgical phase and assess the models' capabilities of assigning the correct role to each track.

\subsubsection{Findings}
\autoref{tbl:abl_role_prediction} presents the accuracy comparison between the scene graph-based approach \cite{ozsoy20224d} and our re-identification-based method across different roles and surgical phases. 
Our re-identification approach achieves notably higher accuracy in every category.
Especially for the circulating nurse and anesthetist, the scene-graph approach underperforms by a large margin compared to our re-identification approach, consistently achieving ~30\% higher accuracy.

\subsection{Experiment 6: Ablation Over Metric Crop and SVM}
\label{subsec:results_ablation_over_metric_crop_and_svm}

\begin{table*}[ht]
\centering
\caption{Inter- and intra-dataset performance (in percentage macro accuracy) of our proposed method in addition to two ablations, \textit{MetricCrop}, and using an \textit{SVM-Gallery}. Columns indicate the training dataset and rows the ablation configuration for a given test dataset. We use the same 4-fold cross-validation splits as in previous experiments. Mean, standard error (using $\pm$ to separate values), and statistical significance with respect to the best performing method are reported for each other method (``***": $p < 0.001$, ``**": $p < 0.01$, ``*": $p < 0.05$, ``ns'': not significant).
}
\label{tbl:al_vs_fixed}
\begin{tabular}{cccclll}
 & & & & \multicolumn{3}{c}{TRAIN} \\
\cmidrule(lr){5-7}
 & & \textit{MetricCrop} & \textit{SVM-Gallery} & SUSTech1K\cite{lidargait} & 4D-OR\cite{ozsoy20224d} & \ordataset \\
\midrule
\multirow{8}{*}{\rotatebox[origin=c]{90}{TEST}}
& \multicolumn{1}{|c|}{\multirow{4}{*}{4D-OR\cite{ozsoy20224d}}} 
  & $\times$    & $\times$    & $65.30 \pm 1.70^{**}$ & $92.00 \pm 0.75^{*}$  & $74.07 \pm 1.81^{**}$ \\
& \multicolumn{1}{|c|}{} 
  & $\times$    & $\checkmark$& $70.39 \pm 1.76^{**}$ & $92.74 \pm 0.50^{*}$  & $79.52 \pm 1.56^{*}$  \\
& \multicolumn{1}{|c|}{} 
  & $\checkmark$& $\times$    & $82.92 \pm 0.94^{**}$ & $95.70 \pm 0.36^{\text{ns}}$       & $85.57 \pm 0.67^{*}$  \\
& \multicolumn{1}{|c|}{} 
  & $\checkmark$& $\checkmark$& $\mathbf{90.06 \pm 0.26}$ & $\mathbf{95.95 \pm 0.45}$ & $\mathbf{89.43 \pm 0.33}$ \\
\cmidrule(lr){2-7}
& \multicolumn{1}{|c|}{\multirow{4}{*}{\ordataset}} 
  & $\times$    & $\times$    & $60.07 \pm 0.60^{**}$ & $54.72 \pm 1.14^{**}$ & $83.02 \pm 2.30^{\text{ns}}$      \\
& \multicolumn{1}{|c|}{} 
  & $\times$    & $\checkmark$& $67.86 \pm 0.87^{*}$  & $62.21 \pm 0.10^{*}$  & $84.48 \pm 1.35^{\text{ns}}$      \\
& \multicolumn{1}{|c|}{} 
  & $\checkmark$& $\times$    & $68.17 \pm 0.76^{*}$  & $58.25 \pm 0.62^{**}$ & $83.59 \pm 2.19^{\text{ns}}$      \\
& \multicolumn{1}{|c|}{} 
  & $\checkmark$& $\checkmark$& $\mathbf{75.78 \pm 0.63}$ & $\mathbf{65.93 \pm 0.64}$ & $\mathbf{85.74 \pm 0.64}$ \\
\bottomrule
\end{tabular}
\end{table*}

\subsubsection{Setup}
To quantify the impact of \textit{\textit{SVM-Gallery}} and \textit{MetricCrop} on model performance, we perform ablation experiments on the three datasets in \autoref{subsubsec:datasets}, using the same four-fold cross-validation.
When ablating \textit{MetricCrop}, we instead follow LidarGait's \cite{lidargait} approach of cropping each individual to fill a virtually rendered point cloud map (see appendix for qualitative examples). 
For the \textit{SVM-Gallery} ablation, rather than training a support-vector machine classifier on the ten gallery sequences, we employ a simple nearest-neighbor search using Euclidean distance at inference time, which is commonly adopted in the re-identification literature \cite{reid_review}.

\subsubsection{Findings}

In \autoref{tbl:al_vs_fixed}, we present the rank-1 macro accuracy with various setups.
Our ablation studies demonstrate that both \textit{MetricCrop} and \textit{SVM-Gallery} components contribute meaningfully to the system's performance.
For intra-dataset scenarios, both components show modest but consistent improvements: accuracy increases by 2\%-3\% for \ordataset and 2\%-6\% for 4D-OR.
The most notable impact appears in inter-dataset evaluations, where these components significantly enhance the model's generalization capabilities ($p < 0.05)$, yielding substantial improvements of up to 18\% in accuracy.

\section{Discussion}

\subsection{Summary and General Observations}

Our empirical findings support several key advances in modeling OR personnel. 
Experiment 1 (\autoref{subsec:results_deficits_of_rgb_based_person_re-identification}) demonstrates that RGB-based models exhibit systematic biases toward OR-specific visual characteristics. 
While these biases manifest as simulation artifacts like streetwear in the 4D-OR dataset, the absence of such distinctive visual cues significantly impairs performance in realistic clinical settings (see intra-dataset \ordataset \autoref{subsec:results_experiment_generalization}). 
In these more challenging environments, models leveraging 3D geometric information offer superior performance.

The generalization experiments further validate these findings - RGB-based methods demonstrate poor transfer to novel environments due to overfitting on domain-specific features. 
However, utilizing depth information as an input modality effectively addresses this limitation. 
Depth-based approaches can attend to fundamental shape and motion characteristics that provide robust identification signals generalizable across environments.

We demonstrate the practical value of our re-identification framework for OR workflow analysis through two key applications. 
First, integrating re-identification capabilities with existing tracking systems enables recovery from missed detections, improving tracking accuracy by over 50\% (Avg. Micro/Macro) during extended surgical procedures. 
Second, direct identity prediction proves substantially more reliable than inferring identity through person-object interactions as implemented in prior work \cite{ozsoy20224d}. 
This highlights a critical limitation of current OR modeling approaches emphasizing role-based analysis over unique identity.
Our qualitative results reinforce these findings - the role-based 3D activity imprints in \autoref{fig:const_heatmaps} reveal significant variations in positioning and behavior patterns even among staff sharing the same surgical role. 
As workflow models expand to encompass multiple operating rooms, clinics, and personnel, reliably modeling the macroscopic behavior of individuals becomes increasingly important.

\subsection{Discussion 1: Deficits of RGB-Based Person Re-Identification}
\label{subsec:discussion_deficits_of_rgb_based_person_re_identification}

Closer inspection of the saliency maps in \autoref{fig:saliency_maps} suggests that 4D-OR\cite{ozsoy20224d} contains distinctive visual cues that ``leak'' identity information, such as street shoes and glasses unique to an individual.
Consequently, CNNs can create spurious correlations, hyper-fixating on these areas.
Such overfitting is particularly evident in the second row in the middle \autoref{fig:saliency_maps} (4D-OR), where despite the target individual pulling the patient table, the saliency maps highlight the socks of the patient, suggesting these influence model decisions more than the stature of the target individual. 
Such clear giveaways can prevent the learning of robust feature representations, which also affects generalization, as seen in \autoref{subsec:results_experiment_generalization}. 

In contrast, the more dispersed saliency patterns observed in \ordataset suggest that CNNs cannot identify such features in more realistic OR settings.
The standardized surgical attire, including scrubs, masks, and protective equipment, effectively removes many distinguishing characteristics that RGB-based methods can learn to exploit.
The observations demonstrate that \ordataset provides a more effective and reliable benchmark for evaluating person re-identification systems intended for real OR environments.

\subsection{Discussion 2: Human-Pose-Based Tracking and Re-Identification}
\label{subsec:discussion_huamn_pose_based_tracking_and_re_identification}

The poor performance of the 3D human-pose-based methods in \autoref{tab:simple_tracking_small} suggests that these approaches require ideal environments to be effective.
In ideal scenarios, naive tracking successfully re-associates lost tracks due to, e.g., occlusions, as reflected in the elevated macro accuracy.
However, multiple individuals working in close proximity can lead to swapped tracks that both the simple heuristic and the KSP tracker \cite{KSP_tracker} cannot cope with.
In the example of \textit{P\_02} and \textit{P\_17} (\autoref{fig:joints_3d_figure}), missed 3D human-pose estimations result in an incorrect re-association.
Without re-identification, such track-swapping inevitably leads to large error accumulation proportional to surgical procedure length.

The robustness of using re-identification to re-assign lost tracks is apparent by the $>$ 50\% increase in detection accuracy for the length of the surgery.
Even with a single annotated sequence in the gallery, this significantly improves pose-based approaches with continued improvements apparent for larger gallery sizes (\autoref{tab:simple_tracking_small}).

\subsection{Discussion 3: Intra and Inter-Domain Performance}
\label{subsec:discussion_intra_inter_domain_performance}
RGB-based methods exhibit near-perfect performance in the intra-domain setting on 4D-OR.
As discussed in \autoref{subsec:discussion_deficits_of_rgb_based_person_re_identification}, this can be attributed to visual artifacts resulting in overfitting on spurious correlations. 
The high performance of point cloud-based methods on 4D-OR can be similarly explained by the limited number of individuals and complexity (five distinct individuals see \autoref{tbl:datasets_overview}).
On \ordataset, when faced with standardized OR attire that removes distinctive visual cues, RGB-based approaches struggle to maintain their performance.
In contrast, the point cloud-based methods maintain high accuracy, suggesting they capture more fundamental and persistent features related to body shape and motion patterns based on 3D geometry.

The inter-domain setting depicts substantial differences in performance between \colorbox{depthexp!15}{Ours (PC)} and other methods.
\colorbox{rgbexp!15}{Ours (RGB)} exhibits significant performance drops (p < 0.01), indicating potential overfitting to dataset-specific characteristics. 
This trend is particularly evident when we train \colorbox{rgbexp!15}{Ours (RGB)} on the small-scale 4D-OR dataset and test on \ordataset, showing a marked decline in performance, validating our hypothesis that RGB generalization in cross-dataset settings is limited.
\colorbox{rgbexp!15}{PAT} achieves better performance than \colorbox{rgbexp!15}{Ours (RGB)} through a powerful architecture that mines clothes-invariant local similarities albeit at a much higher computational complexity (10x the parameter count).

In contrast, when tested on out-of-domain datasets, \colorbox{depthexp!15}{Ours (PC)} maintains a higher accuracy compared to all other methods (p < 0.01), indicating that it learns robust features for generalization across diverse domains.
While this robustness partly stems from the richer structural information encoded in point cloud data, \textit{MetricCrop} and \textit{SVM-Gallery} contribute to performance gains as \colorbox{depthexp!15}{LidarGait} fails to perform better than \colorbox{rgbexp!15}{PAT}, despite using point cloud data.
These components significantly enhance the \colorbox{depthexp!15}{Ours (PC)}'s ability to generalize across variations in shape and motion in person sequences (see ablations \autoref{subsec:results_ablation_over_metric_crop_and_svm}).

\subsubsection{Latent Space Visualization}

T-SNE \cite{van2008visualizing} projection of the latent spaces (see \autoref{fig:qual-results}) depict different latent space arrangements, highlighting differences between the two modalities.
We observe appearance-based clustering similarity in the RGB latent space, particularly regarding surgical attire. 
As such, \textit{person 2} and \textit{person 8} are grouped together due to their similar clothing. 
However, as sterility requirements determine OR attire, the same roles often dress similarly.
Consequently, relying on attire to re-identify individuals primarily recovers an individual's role instead of their identity.

In contrast, the point cloud latent space appears better separated, with distinct clusters for short and tall individuals and unique shape characteristics such as the helmets worn by surgeons.
This leads to a better distinction over the individual beyond what RGB features can achieve.

\subsection{Discussion 4: Downstream Analysis of OR Workflows through 3D Activity Imprints}
\label{subsec:discussion_downstream_analysis_of_OR_workflows_through_heat_maps}

Our generated 3D activity imprints reveal several important insights about movement patterns and team dynamics, with implications for workflow optimization and OR design.

\subsubsection{Individual Adaptations and Preferences.}

The consistent side preferences (green surgeon favoring the left side, pink surgeon the right) observed in \autoref{fig:heatmaps} suggest that the head surgeons develop positional preferences that persist across multiple procedures. 
Moreover, these preferences appear to influence the team's spatial organization.
The pink scrub nurse, for instance, occupies different locations depending on which head surgeon is leading the operation (surgery 1\&2 vs. 3\&4).

\subsubsection{Team Configurations and Space Utilization.}

In \autoref{fig:const_heatmaps}, the localized activity clusters of constellation 2 suggest an optimized spatial arrangement that minimizes unnecessary movement and potential conflicts. 
In contrast, the more dispersed patterns in constellation 1 indicate less efficient spatial organizations. 
The presence of two high-activity centers for the assistant surgeon in constellation 3, separated by the head surgeon, suggests a potentially suboptimal workflow that may require unnecessary movement and coordination overhead.

\subsection{Discussion 5: Downstream Analysis of Role Prediction via Person Re-Identification}
\label{subsec:discussion_downstream_analysis_of_role_prediction_via_person_re_identification}

The scene graph approach's varying performance across different roles highlights the limitations of relying solely on contextual interactions for role prediction. 
This method struggles with roles that have less consistent interaction patterns, such as the circulating nurse and anesthetist. 
The fundamental challenge lies in the dependence on observing role-specific actions or interactions, which may not always be present or visible during a procedure phase. 
When individuals temporarily deviate from their typical role-specific behaviors or when interactions are occluded or subtle, purely contextual approaches can fail to maintain accurate role assignments.

Our re-identification-based approach demonstrates the benefits of focusing on identity rather than contextual cues. 
This method remains robust by identifying the individual and then mapping to their assigned role even when staff members are not actively engaged in role-typical behaviors. 
While this approach requires a pre-existing mapping between individuals and their roles, this information is usually easily obtained. 
Furthermore, our method could be combined with role-based prediction methods, incorporating further context and interactions to improve predictions.

\subsection{Discussion 6: Ablation Over Metric Crop and SVM}
\label{subsec:discussion_ablation_over_metric_crop_and_svm}

The \textit{MetricCrop} component substantially improves our method's generalization by preserving absolute scale information, enabling more reliable comparisons of individuals across different environments. 
In contrast, without \textit{MetricCrop}, the network tends to learn dataset-specific relative scales, which fail to transfer effectively when encountering new domains with varying individual sizes and segmentation quality.

The \textit{SVM-Gallery} component effectively structures the latent space for new domains. 
While simple nearest-neighbor matching suffices for intra-dataset scenarios, the SVM's ability to learn decision boundaries from multiple gallery instances becomes crucial when generalizing to new environments. 
The performance improvements with \textit{SVM-Gallery} suggest effectiveness in capturing and adapting to distribution shifts between different environments, utilizing OR-specific domain characteristics.

\subsection{Limitations and Ethical Considerations}

Data acquisitions, and thus, publically available OR datasets, are limited due to anonymization protocols and privacy regulations in healthcare \cite{bastian2023disguisor,maier2022surgical}.
A key advantage of depth-based approaches, beyond robustness, is privacy protection. 
While sharing RGB operating room data poses significant privacy risks \cite{bastian2023disguisor}, point cloud data naturally obscures identifying details while preserving essential shape and motion patterns. 
This privacy-preserving characteristic could enable secure data sharing between medical institutions to develop more robust models through multi-center collaboration.

We perform experiments on two available OR datasets: \ordataset, comprising 13 identities, and 4D-OR \cite{ozsoy20224d}, comprising 5 identities.
Nevertheless, further evaluation with datasets containing more individuals will help to confirm our findings and illuminate additional challenges when integrating our methods into workflow analysis systems.
Additionally, the datasets are based on simulated conditions, which may not fully reflect real-world OR variability. 
A further challenge emerges when multiple individuals in the OR share similar physical characteristics, such as stature and body size, making it more difficult to differentiate between them.
Here, motion and action disentanglement could play a larger role in predicting their identity, or for example, incorporating additional contextual events such as with scene graphs \cite{ozsoy20224d}.

Our approach is not yet fully end-to-end as it requires pre-segmented 3D point cloud sequences for each individual, which are obtained with a separately trained weakly-supervised segmentation approach \cite{bastian2023segmentor}. 
In the future, methods could detect, segment, and re-identify all unique individuals in a scene simultaneously, reducing erroneous predictions and improving the association of identities to each individual.

In addition to these technical limitations, deploying tracking systems in the OR raises important ethical considerations \cite{morris2024current}.
While the field is motivated to build technologies to enhance surgical efficiency and improve patient outcomes, there is a potential risk of their misuse, such as being employed to monitor healthcare workers in punitive rather than supportive ways. 
To minimize these risks, clinics must establish clear ethical guidelines that dictate how tracking and performance data can be used.
This continues to be an active discussion area given the deployment of systems such as the OR Black-Box \cite{jung2020first} and other recording devices in ORs.

\section{Conclusion}

Our work addresses an existing gap in surgical domain modeling, namely that OR personnel are only modeled by their role while more nuanced individual characteristics are ignored.
We propose a method for personnel re-identification in operating rooms (ORs) that is robust to different environments and large domain changes which addresses both short- and long-term tracking and identification challenges.
Our methodological design is motivated by insights into the challenges of modeling personnel movement and identities in ORs. 
Due to large domain shifts between different clinics, RGB-based models tend to overfit to distinct visual cues, leading to performance degradation.

To address these shortcomings, we demonstrate the superior performance and generalizability of our proposed 3D point cloud sequence-based tracking model that effectively captures body shape and articulated motion patterns over potentially misleading visual appearance.
Finally, we showcase the practical applications of our approach for workflow analysis and optimization through 3D activity imprint generation to aid in the analysis of OR teams. 
By introducing a non-invasive and robust method for identifying surgical personnel, our approach shifts the analysis of surgical procedures to a more individualized level.
This lays the foundation for more adaptive and tailored AI-driven systems in surgical settings, facilitating individualized feedback, optimizing team coordination, reducing inefficiencies, and ultimately enhancing patient safety and care.

\section*{CRediT Authorship Contribution Statement}

\textbf{Tony Danjun Wang}: Writing – Review \& Editing, Writing – Original Draft, Software, Methodology, Validation, Investigation, Visualization, Data Curation.
\textbf{Lennart Bastian}: Conceptualization, Methodology, Validation, Writing – Review \& Editing, Writing – Original Draft, Project administration.
\textbf{Tobias Czempiel}: Conceptualization, Writing – Review \& Editing, Writing – Original Draft.
\textbf{Christian Heiliger}: Writing – Review \& Editing, Funding acquisition. 
\textbf{Nassir Navab}: Writing – Review \& Editing, Supervision, Resources, Funding acquisition.

\section*{Declaration of Competing Interest}

The authors declare that they have no known competing financial interests or personal relationships that could have appeared to influence the work reported in this paper.

\section*{Data and Code Availability}

The 4D-OR dataset, as well as the SUSTech1K dataset, are publicly available.
We publish our annotations of \ordataset, which is part of the MM-OR dataset by Özsoy et al. \cite{ozsoy2025mmor}, making all experiments reproducible.
Our code will be available at \href{https://github.com/wngTn/or_reid}{https://github.com/wngTn/or\_reid}

\section*{Acknowledgement}

This work was partly supported by the state of Bavaria through Bayerische Forschungsstiftung (BFS) under Grant AZ-1592-23-ForNeRo and the German Federal Ministry for Economic Affairs and Climate Action (BMWK) through the Central Innovation Programme for SMEs (ZIM) under Grant KK 5389102BA3.

\bibliographystyle{elsarticle-num} 
\bibliography{refs_REBUTTAL}

\clearpage
\pagebreak
\appendix
\label{appendix}
\begin{strip} %
  \centering
  {\Large Appendix: Beyond Role-Based Surgical Domain Modeling}
\end{strip}

\noindent
In our appendix, we provide additional experiments in \ref{app_sec:additional_experiments}, demonstrating a method to fuse RGB and depth modality, and detailed performance analysis through confusion matrices. 
In \ref{app_sec:ablation_experiments}, we present several ablation studies that further detail the design choices of our method.
Finally, in \ref{app_sec:additional_figures} we depict additional qualitative examples and tables regarding our implementation details.

\section{Additional Experiments}
\label{app_sec:additional_experiments}

\subsection{Modality Feature Fusion}

To better understand the role of feature fusion strategies in our setting, we design a simple RGB-Depth fusion strategy by taking a forward pass through both the RGB and Depth encoding strategies detailed in \ref{subsec:method_inference}, pooling the embeddings from both modalities. 
Our multi-view voting strategy is then carried out over the pooled features from both modalities when determining the closest fit from the gallery.

The results can be seen in \ref{tab:fusion_performance}. 
Notably, this feature fusion strategy improves results for the intra-domain (first two rows) cases as both modalities perform competitively when trained and tested on the same data. 
For the inter-domain generalization cases (last four rows), the results are mixed with neither modality consistently better. 
In some cases, the significantly worse RGB features adversely affect the fusion results, leading to a notable performance decrease.

\begin{table}[htbp]
    \centering
    \normalsize
    \resizebox{\linewidth}{!}{
    \begin{tabular}{ll| lll}
    \toprule
    Train & Test & Fusion & Depth & RGB \\
    \midrule
    4D\_OR & 4D\_OR & \textbf{98.46 $\pm$ 0.10} & 97.84 $\pm$ 0.16$^{**}$ & 98.18 $\pm$ 0.37$^{\text{ns}}$ \\
    OR\_ReID\_13 & OR\_ReID\_13 & \textbf{92.62 $\pm$ 1.17} & 91.58 $\pm$ 1.33$^{\text{ns}}$ & 81.72 $\pm$ 2.23$^{**}$ \\
    OR\_ReID\_13 & 4D\_OR & 92.87 $\pm$ 0.33$^{**}$ & \textbf{94.09 $\pm$ 0.23} & 73.02 $\pm$ 1.47$^{***}$ \\
    4D\_OR & OR\_ReID\_13 & 76.16 $\pm$ 0.80$^{*}$ & \textbf{77.65 $\pm$ 0.45} & 51.53 $\pm$ 2.06$^{***}$ \\
    SUSTech1K & 4D\_OR & 93.98 $\pm$ 0.18$^{\text{ns}}$ & \textbf{94.13 $\pm$ 0.27} & 77.70 $\pm$ 0.76$^{***}$ \\
    SUSTech1K & OR\_ReID\_13 & \textbf{84.24 $\pm$ 0.63} & 83.40 $\pm$ 0.73$^{\text{ns}}$ & 70.84 $\pm$ 0.80$^{***}$ \\
    \bottomrule
    \end{tabular}
    }
    \caption{Performance comparison across different training and testing configurations. Values represent mAP scores ($\%$) with standard deviations. The best value for each metric is highlighted in \textbf{bold}. 
    Statistical significance (paired t-test) with respect to the best performing method is shown for each other method (***: $p < 0.001$, **: $p < 0.01$, *: $p < 0.05$, ``ns'': not significant).}
    \label{tab:fusion_performance}
\end{table}

\subsection{Confusion Matrices}

\begin{figure*}[h!t]
    \centering
    \includegraphics[width=\textwidth]{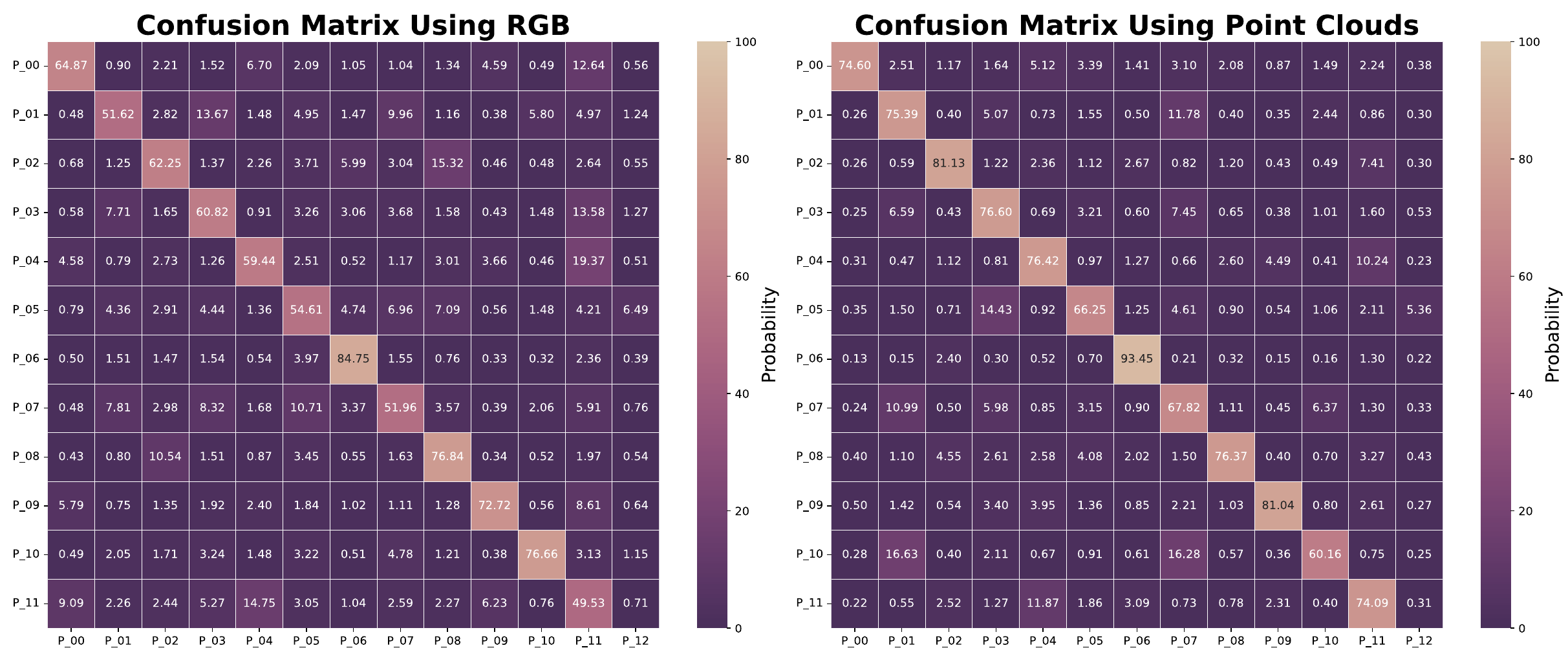}
    \caption{Confusion matrix over probe identities (y-axis) and gallery identities (x-axis) of our method using \colorbox{rgbexp!15}{RGB} (left) and \colorbox{depthexp!15}{point cloud} (right) as input. The probabilities are taken from the probability vectors acquired from the SVM classifier (see \autoref{subsec:method_inference}). $P\_12$ only appeared briefly throughout the dataset and was, therefore, not included in the probe set of this split.
    }
    \label{fig:conf_matrix}
\end{figure*}

Figure \ref{fig:conf_matrix} depicts confusion matrices for a single split of Ours (RGB) and Ours (PC). 
Notably, both modalities excel in distinguishing \textit{P\_06}, despite their infrequent appearance in the dataset.
This accuracy is largely because of \textit{P\_06}'s distinct visual characteristics, such as the absence of head coverings and a partially shaved head, as illustrated in \ref{fig:data_overview}).
Additionally, the lack of gowns and surgical caps results in a unique body shape and articulated motion.

However, when using RGB input, the model struggles to differentiate \textit{P\_01}, assigning a lower probability (51.62\%) compared to the higher confidence achieved with point clouds (75.39\%). For \textit{P\_01}, the most significant confusion in the RGB input occurs with \textit{P\_03} (13.67\%), likely due to their similar attire—scrubs, face masks, shoes, and bouffant hats—resulting in nearly identical appearances (see images in \ref{fig:data_overview})
Similarly, the RGB input encounters difficulties in distinguishing between \textit{P\_04} and \textit{P\_11} (19.37\% confusion), both dressed in surgeons' attire with minimal visual variation, as depicted in \ref{fig:data_overview}. On the other hand, the point cloud method, less dependent on texture, relies on subtle differences in shape and motion to achieve more accurate differentiation between these two surgeons (10.24\% confusion).
In contrast, the most significant confusion for our point cloud method arises with \textit{P\_07} (11.78\%), likely due to a comparable stature and body size. This highlights the distinct features emphasized by each modality: RGB focuses on visual appearance linked to surgical roles, while point clouds prioritize shape and geometry.

\section{Ablation Experiments}
\label{app_sec:ablation_experiments}

\subsection{Results 1: Ablation Over Pre-Segmentation of Images}

\subsubsection{Setup}
\begin{figure*}[ht]
    \centering
    \caption{Semi-autonomous SAM2 \cite{ravi2024sam2} segmentations to crop individuals as a pre-processing step for re-identification.}
    \includegraphics[width=0.8\textwidth]{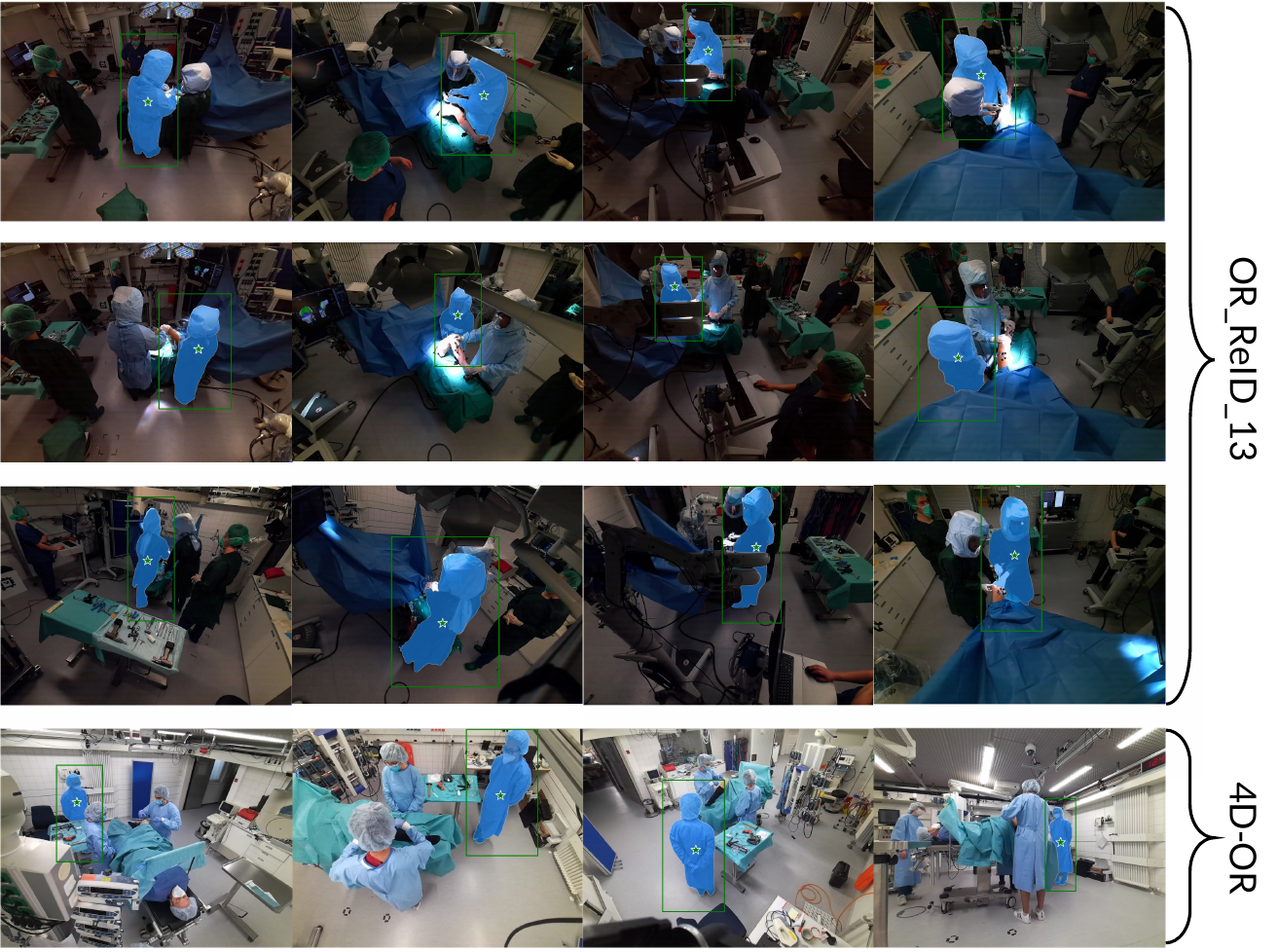}
    \label{fig:sam2_backgrounds}
\end{figure*}

RGB images contain much more background noise compared to our segmented 4D point clouds. 
Thus, we validate whether shape and motion cues can be better extracted from RGB images if we segment them first to remove unnecessary and potentially confusing background information.
To obtain 2D segmentation masks, we project the person bounding box and a query point into 2D per individual and frame using the previously 3D segmented point clouds. 
We then use SAM2 \cite{ravi2024sam2} to obtain 2D video segmentation for each person using the projected 2D bounding boxes and query points as input (see qualitative examples in figure \ref{fig:sam2_backgrounds}).

\subsubsection{Findings}

In table \ref{tbl:rgb_mask}, we depict the performance of segmented RGB images vs. unsegmented RGB images in Acc. (Macro).
For intra-dataset scenarios, segmentation consistently improves performance.
On 4D-OR, segmentation yields a marginal improvement (96.33\% vs. 96.25\%), while on \ordataset, the improvement is more substantial (78.55\% vs. 73.23\%).

In inter-dataset scenarios, the results are more mixed.
When training on 4D-OR and testing on \ordataset, segmentation provides a notable advantage (51.19\% vs. 36.52\%). 
However, when training and testing the other way around, unsegmented RGB performs better (52.69\% vs. 52.28\%).
Similar mixed results are observed with SUSTech1K as a training dataset.
While segmentation slightly improves the performance when testing on 4D-OR (63.03\% vs. 62.59\%), it underperforms compared to unsegmented RGB images when testing on \ordataset (55.66\% vs. 60.42\%)

\begin{table}[ht]
\centering
\normalsize
\setlength{\tabcolsep}{4pt}  %
\renewcommand{\arraystretch}{1.1} %
\caption{
Performance in percentage macro accuracy between using segmented RGB images as input and unsegmented RGB images.
We use the same 4-fold cross-validation splits as in previous experiments.
Statistical significance with respect to the best performing method is shown for each other method (***": $p < 0.001$, **": $p < 0.01$, *": $p < 0.05$, ns": not significant).
}
\label{tbl:rgb_mask}
\resizebox{\columnwidth}{!}{
\begin{tabular}{ccll}
\toprule
\multicolumn{1}{l|}{\textbf{Train}} &
  \multicolumn{1}{l|}{\textbf{Test}} &
  \textbf{RGB-Segmented} & \textbf{RGB} \\\hline \hline
\multicolumn{4}{l}{\multirow{2}{*}{Intra Dataset}} \\ & & & \\\hline \hline
\multicolumn{2}{c|}{\multirow{1}{*}{4D-OR}} 
    & $\mathbf{96.33\pm 0.93}$ & $96.25\pm0.70^{\text{ns}}$\\\hline
\multicolumn{2}{c|}{\multirow{1}{*}{\ordataset}} 
    & $\mathbf{78.55 \pm 2.72}$ & $73.23 \pm3.13^{\text{ns}}$\\\hline \hline
\multicolumn{4}{l}{\multirow{2}{*}{Inter Dataset}} \\ & & & \\\hline \hline
\multicolumn{1}{c|}{\ordataset} & \multicolumn{1}{c|}{4D-OR} & $52.28 \pm 1.18^{\text{ns}}$ & $\mathbf{54.69 \pm2.69}$\\\hline
\multicolumn{1}{c|}{4D-OR} & \multicolumn{1}{c|}{\ordataset} & $\mathbf{41.19 \pm 2.67}$ & $36.52 \pm2.38^{\text{ns}}$ \\\hline
\multicolumn{1}{c|}{SUSTech1K} & \multicolumn{1}{c|}{4D-OR} & $\mathbf{63.03 \pm 1.27} $ & $62.59 \pm 2.17^{\text{ns}}$\\\hline
\multicolumn{1}{c|}{SUSTech1K} & \multicolumn{1}{c|}{\ordataset}  & $55.66 \pm 1.74^{\text{ns}}$ & $\mathbf{60.42 \pm1.05}$\\\bottomrule
\end{tabular}
}
\end{table}

\subsection{Results 2: Ablation over Maximum Number of Frames in a Sequence}

\begin{figure*}[h]
    \centering
    \includegraphics[width=\textwidth]{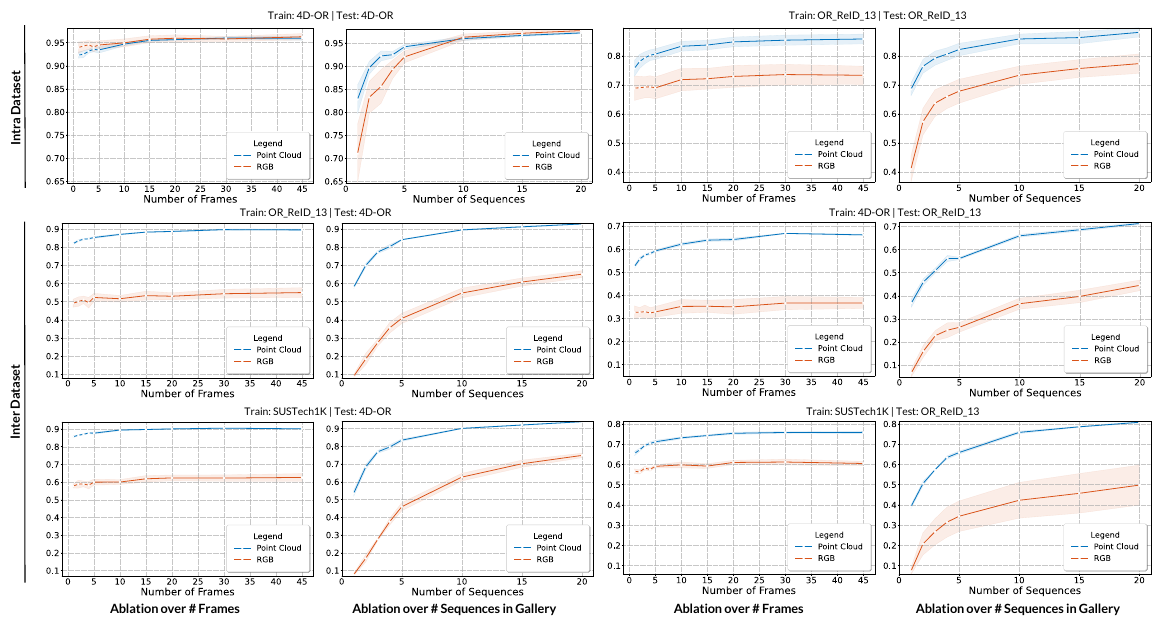}
    \caption{
    Ablation over the number of frames (columns 0 and 2) used during inference and the number of sequences (columns 1 and 3) in the gallery. 
    We perform the ablations on \ordataset with a four-fold cross-validation.
    For both ablations, we report the rank-1 macro accuracy of our method for \colorbox{rgbexp!15}{RGB} input and \colorbox{depthexp!15}{point cloud} input and their corresponding standard error.
    }
    \label{fig:ablation_plots}
\end{figure*}

\subsubsection{Setup}
To evaluate the performance impact of the number of frames in a sequence, we vary the maximum number of frames used during inference. 
We maintain consistency with our previous evaluation protocol by applying the same cross-validation procedure described in section \ref{subsec:results_experiment_generalization} for both intra- and inter-dataset scenarios. 
The analysis encompasses both our point cloud sequence method and RGB sequence input variants.

\subsubsection{Findings}
\ref{fig:ablation_plots} (1st and 3rd column) reveals distinct patterns in how the sequence length affects recognition performance across different modalities. For point cloud input, increasing the number of frames yields substantial improvements, particularly when testing on \ordataset, where accuracy increases by approximately 10\% when extending from 1 to 10 frames. 
This improvement continues beyond 10 frames, showing an additional 3\% gain in \ordataset intra-dataset evaluation.
RGB input demonstrates more modest improvements, with only about 3\% accuracy gain when increasing from 1 to 10 frames during \ordataset testing. Beyond 10 frames, performance improvements generally diminish across most scenarios.

Testing on 4D-OR \cite{ozsoy20224d} shows different characteristics, with only marginal accuracy improvements as frame count increases, and performance consistently plateauing after 10 frames.

\subsection{Results 3: Ablation over Number of Sequences in a Gallery}
\label{subsec:results_ablation_over_number_of_sequences_in_gallery}

\subsubsection{Setup}
To understand how the gallery size affects re-identification performance, we conduct experiments varying the number of sequences per person in the gallery. 
We maintain consistency with our previous evaluation protocol by applying the same cross-validation procedure described in section \ref{subsec:results_experiment_generalization} for both intra- and inter-dataset scenarios. 
The analysis encompasses both our point cloud sequence method and RGB sequence input variants.

\subsubsection{Findings}
Figure \ref{fig:ablation_plots} (2nd and 4th column) reveals a substantial performance gap between point cloud and RGB modalities, particularly when the gallery contains fewer sequences. 
This disparity is most pronounced in the intra-dataset evaluation on \ordataset, where using a single gallery sequence results in a 28\% accuracy difference favoring point cloud input over RGB.

The performance gap becomes even more pronounced in inter-dataset scenarios, particularly when training on SUSTech1K and testing on either \ordataset or 4D-OR. In these cases, the RGB modality's performance deteriorates severely, with accuracy dropping below 1\% when limited to a single gallery sequence. 

Furthermore, in intra-dataset scenarios, the performance plateaus after 10 sequences, suggesting that additional gallery annotations yield diminishing returns. 
Conversely, in inter-dataset settings, performance continues to improve even beyond 10 sequences.

\subsection{Results 4: Ablation Over Encoder Size}
\label{subsec:results_ablation_over_encoder_size}

\subsubsection{Setup}
To better understand the impact of backbone complexity, we conduct experiments comparing ResNet-9, ResNet-50, and ResNet-101.
We maintain consistency with our previous evaluation protocol by applying the same cross-validation procedure described in \ref{subsec:results_experiment_generalization} for both intra- and inter-dataset scenarios. 
The analysis encompasses our point cloud sequence method with different encoder backbones.

\subsubsection{Findings}

Table \ref{tab:backbone_performance} shows that ResNet-9 consistently achieves the highest accuracy across all evaluation settings. 
For intra-dataset evaluation, ResNet-9 achieves 97.84\% on 4D\_OR while ResNet-50 and ResNet-101 show decreased performance at 94.88\% and 92.62\%, respectively. 
A similar pattern is observed on \ordataset, where accuracy drops from 91.58\% with ResNet-9 to 82.52\% with ResNet-50, and further to 75.82\% with ResNet-101.

The performance degradation with deeper networks is more pronounced in inter-dataset scenarios. When training on \ordataset and testing on 4D\_OR, accuracy drops from 94.09\% with ResNet-9 to 84.23\% with ResNet-50 and 77.41\% with ResNet-101. 
The reverse setting shows an even larger gap, with ResNet-9's 77.65\% accuracy dropping to 57.65\% for ResNet-50 and 47.21\% for ResNet-101. 
Similarly, when training on SUSTech1K, performance on both test sets decreases as the network's depth increases, with ResNet-9 outperforming deeper architectures by 9-15 percentage points.

\begin{table}[]
    \centering
    \caption{Performance in percentage macro accuracy between Ours (PC) using ResNet-9, ResNet-50, and ResNet-101 as backbone. Mean and standard errors are reported. Statistical significance (paired t-test) with respect to the best performing method is shown for each other method (***: $p < 0.001$, **: $p < 0.01$, *: $p < 0.05$, ns'': not significant).}
    \label{tab:backbone_performance}
    \resizebox{\columnwidth}{!}{
    \begin{tabular}{ll| lll}
    \toprule
    Train & Test & ResNet-9 & ResNet-50 & ResNet-101 \\
    \midrule
    4D\_OR & 4D\_OR & 
    \textbf{97.84 $\pm$ 0.16} & $94.88 \pm 0.74 ^{*}$ & $92.62 \pm 1.02^{*}$ \\
    OR\_ReID\_13 & OR\_ReID\_13 & 
    $\mathbf{91.58 \pm 1.33}$ & $82.52 \pm 2.14^{*}$ & $75.82 \pm 2.62^{*}$ \\
    OR\_ReID\_13 & 4D\_OR & 
    $\mathbf{94.09 \pm 0.23}$ & $84.23 \pm 0.41^{***}$ & $77.41 \pm 0.85^{***}$ \\
    4D\_OR & OR\_ReID\_13 & 
    $\mathbf{77.65 \pm 0.45}$ & $57.65 \pm 1.24^{***}$ & $47.21 \pm 1.68^{***}$ \\
    SUSTech1K & 4D\_OR & 
    $\mathbf{94.13 \pm 0.27}$ & $84.82 \pm 0.79^{**}$ & $83.11 \pm 1.17^{**}$ \\
    SUSTech1K & OR\_ReID\_13 & 
    $\mathbf{83.40 \pm 0.73}$ & $73.08 \pm 0.89^{**}$ & $68.87 \pm 1.13^{**}$ \\
    \bottomrule
    \end{tabular}
    }
\end{table}

\subsection{Discussion 1: Ablation Over Pre-Segmentation of Images}

The explicit shape information through segmentation offers minimal advantages in the intra-dataset scenario on 4D-OR since networks can easily overfit on the dataset's artifacts.
However, the slightly larger gap observed in \ordataset indicates that shape information becomes more valuable in more realistic OR settings where traditional visual cues are less reliable.

The modest improvements and deteriorations from explicit shape segmentation in the inter-dataset scenarios suggest that even when we attempt to direct RGB models toward shape and motion by removing background noise, they still struggle to learn generalizable features.

Moreover, a fundamental limitation of 2D segmentations is their inability to accurately represent shape, as they lack absolute scale information. 
This loss of crucial depth cues makes them inherently less informative than 3D point clouds.

Additionally, since cameras in ORs are often ceiling-mounted \cite{bastian2023disguisor}, they frequently capture people from a top-down perspective, where human shape becomes less distinguishable.

\subsection{Discussion 2: Ablation over Maximum Number of Frames in a Sequence}

Our findings show that the number of frames plays an important role for both modalities, though with different degrees of effectiveness.
For RGB sequences, understanding texture does not require as many frames, as the visual information is typically clearer and less noisy than point clouds (i.e., better signal-to-noise ratio). 
In contrast, point cloud sequences show marked improvements with additional frames, as the accumulation of spatial data helps refine and distinguish individual shapes.
This sequential refinement proves particularly valuable for capturing articulated motion patterns and enhances the model's resilience to outliers, leading to more reliable identification in complex, real-world scenarios.

While increasing frame count generally improves performance, we observe diminishing returns beyond 10 frames. 
This plateau can be partially attributed to our dataset characteristics, where some sequences are naturally limited to 10 frames (see \ref{fig:data_plots}). 
Additionally, when evaluating on the 4D-OR dataset \cite{ozsoy20224d}, we find that even shorter sequences achieve high performance, primarily since 4D-OR contains few subjects that are easily recognizable with minimal temporal information.

\begin{table*}[h]
    \centering
    \caption{Rank-1 Accuracy (in percentage) of each person when tracking individuals over an entire surgery based on associating 3D human poses between frames (Simple Tracking, KSP Tracker) compared to re-identification based tracking [Ours (PC, $n$)], where $n$ denotes the number of sequences per person in the gallery.
    }
    \label{tab:simple_tracking}
    \resizebox{\textwidth}{!}{%
    \begin{tabular}{l|cccccccccc}
        \toprule
        \textbf{Method} & P\_00 & P\_01 & P\_03 & P\_07 & P\_08 & P\_09 & P\_10 & P\_12 & Avg. Micro & Avg. Macro \\
        \midrule
        Simple Tracking \cite{ozsoy20224d} & 5.26 & 9.51 & 83.04 & 22.40 & 16.23 & 7.62 & 28.54 & 91.49 & 15.79 & 33.01 \\
        KSP Tracker \cite{KSP_tracker} & 25.96 & 8.38 & 0.00 & 8.69 & 17.62 & 44.59 & 26.67 & 0.00 & 19.83 & 16.49 \\ \hline
        Ours (PC, $1$) & 81.73 & 68.86 & 60.55 & 20.05 & 52.67 & 62.43 & 34.79 & 87.00 & 55.84 & 58.51 \\
        Ours (PC, $5$) & 86.32 & 75.79 & 73.64 & 31.33 & 89.65 & 81.84 & 35.86 & 100.0 & 73.30 & 71.80 \\
        Ours (PC, $10$) & 87.56 & 82.13 & 82.18 & 37.00 & 91.24 & 87.65 & 34.51 & 100.0 & 76.54 & 75.28 \\
        \bottomrule
    \end{tabular}
    }

\end{table*}

\subsection{Discussion 3: Ablation over Number of Sequences in a Gallery}

Ours (PC) exhibits superior performance compared to Ours (RGB) when the gallery contains fewer sequences, which is largely due to a more effective latent space organization, as visualized in \ref{fig:qual-results}. 
This better-structured embedding space enables more reliable identification even with limited gallery examples.

The continued improvement in inter-domain scenarios beyond 10 sequences highlights the effectiveness of the SVM classifier in leveraging few-shot examples for domain adaptation. The classifier's ability to establish more refined decision boundaries with additional gallery samples suggests that the challenge of inter-domain generalization can be partially mitigated through careful gallery curation, even when the number of available examples is limited.

\subsection{Discussion 4: Ablation Over Encoder Size}

While larger backbones theoretically offer greater capacities for feature extraction, our experiments show that ResNet-9 consistently outperforms both ResNet-50 and ResNet-101 in our specific context. This aligns with findings from GaitBase \cite{fanOpenGaitRevisitingGait2023}, where similar patterns were observed. The superior performance of the lighter backbone is likely
due to the smaller input image dimension (64x64) and dataset size in the re-identification
domain, with larger models being more prone to overfitting. Another factor may be the specialized nature of our task, where simpler architectures could better capture the essential shape and motion features without learning spurious correlations.

\section{Additional Figures}
\label{app_sec:additional_figures}

In table \ref{tab:simple_tracking}, we present the extended results table of the human-pose-based tracking and re-identification experiment of the main paper, featuring the accuracies for each individual.
As discussed in the main paper, the naive tracking method's macro accuracy is higher since it achieves high accuracies on P\_03 (83.04\%) and P\_12 (91.49\%) due to these individuals only once entering and shortly exiting the OR.

In \ref{tbl:param_count}, we compare the number of parameters used in our ResNet-9-based model and the vision transformer-based method PAT proposed by Ni et al. \cite{ni2023part}.

\begin{table}[h]
    \centering
    \caption{Comparison of parameter counts between our model and the vision transformer-based architecture PAT \cite{ni2023part}}
    \begin{tabular}{lcc}
    \toprule
    \textbf{Method} & \textbf{Ours} & \textbf{PAT} \\
    \midrule
    \textbf{\# Parameter} & 7.05M & 86.62M \\
    \bottomrule
    \end{tabular}
    \label{tbl:param_count}
\end{table}

\begin{figure*}[p]
    \centering
    \includegraphics[width=\textwidth]{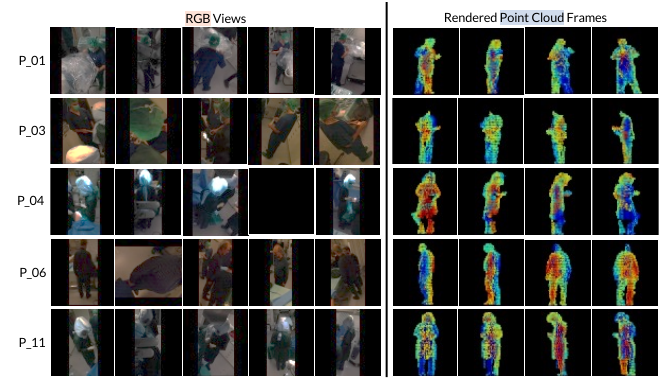}
    \caption{Depiction of a single frame (from a sequence) of five different identities from the \ordataset. A black image in the RGB views denotes an obstructed view. Person 1 (P\_01) looks visually similar to person 3 since they both wear the same scrubs. Person 4 looks visually similar to person 11 since they both wear the same surgical attire. Person 6 stands out visually as they do not wear any head covering and exhibit shaved head parts.}
    \label{fig:data_overview}
\end{figure*}

In \ref{fig:data_overview}, we visualize five identities from our internal \ordataset, along with their corresponding RGB images and rendered point clouds.

\begin{figure*}[p]
    \centering
    \includegraphics[width=.95\textwidth]{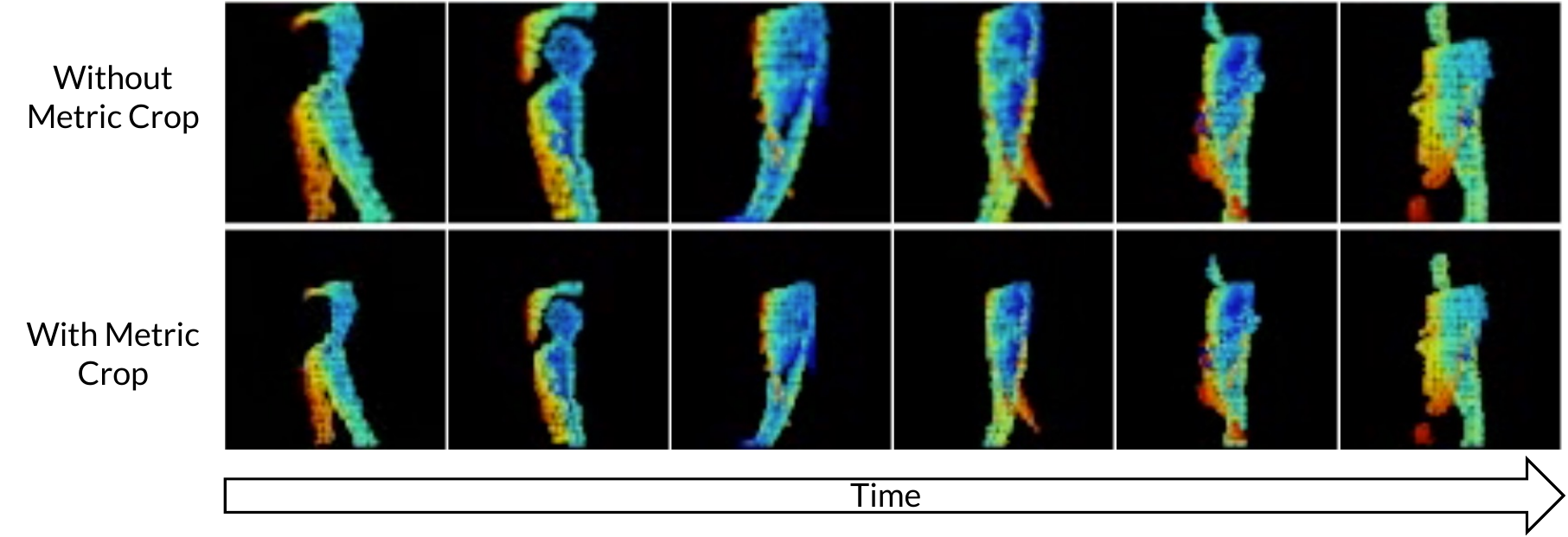}
    \caption{Comparison of rendered point cloud sequences with and without \textit{MetricCrop}. Without \textit{MetricCrop} (following \cite{lidargait}), the rendered point clouds exhibit varying scales due to their incompleteness, while with \textit{MetricCrop}, the absolute scale of the person remains consistent throughout the sequence.
    We sampled every fifth frame of the sequence in this figure.
    }
    \label{fig:metric_crop}
\end{figure*}

\ref{fig:metric_crop} illustrates the impact of using \textit{MetricCrop} on the rendered point clouds. Without \textit{MetricCrop}, the incomplete point clouds exhibit inconsistent scaling across different frames, while the use of \textit{MetricCrop} maintains a consistent scale throughout.

\ref{tab:dataset_splits} depicts the number of sequences for training and testing per split.

\begin{table}[t]
    \normalsize
    \caption{Number of sequences per splits for OR\_ReID\_13 and 4D-OR.}
    \label{tab:dataset_splits}
    \resizebox{\columnwidth}{!}{
    \centering
        \begin{tabular}{lcccc}
            \toprule
            Dataset            & Split 1 & Split 2 & Split 3 & Split 4 \\
            \midrule
            OR\_ReID\_13 Train  & 1874    & 2076    & 2348    & 1801    \\
            OR\_ReID\_13 Test   & 3530    & 3328    & 3056    & 3603    \\
            4D-OR Train        & 1317    & 1560    & 1582    & 1328    \\
            4D-OR Test         & 2402    & 2159    & 2137    & 2391    \\
            \bottomrule
        \end{tabular}%
        }
\end{table}

\ref{tab:da_overview} shows the different data augmentation setups we used for each experiment for each method based on optimal hyperparameters. Horizontal flip (HF) means flipping the entire sequence horizontally, random cropping (RC) crops up to 20\% of the image, local grayscale transformation \cite{gong2021eliminate} converts patches of the input image to grayscale, random erasing \cite{zhong2020random} replaces patches with black pixels, and Gaussian noise adds noise along the colormap used in the point cloud projections. 

\clearpage

\begin{table*}[tp]
\centering
        \caption{Overview of data augmentations used in our experiments. The datasets we use are 4D-OR \cite{ozsoy20224d}, our internal dataset \ordataset, and SUSTech1K \cite{lidargait}. HF is horizontal flipping, RC is random cropping, LGT is local grayscale transformation \cite{gong2021eliminate}, RE is random erasing \cite{zhong2020random}, and GN is Gaussian noise.}
    \label{tab:da_overview}
    \resizebox{0.8\linewidth}{!}{
\begin{tabular}{l|ccccccc}
            \toprule
\multirow{2}{*}{Method}     & \multirow{2}{*}{Train}    & \multirow{2}{*}{Test}     & \multicolumn{5}{c}{Data Augmentations} \\ \cline{4-8} 
                            &                           &                           & HF          & RC         & LGT         & RE & GN \\ \hline
\multirow{6}{*}{Ours (RGB)} & 4D-OR                     & 4D-OR                     & \checkmark  & \checkmark & \checkmark  & & \\
                            & \ordataset                & \ordataset                &             &            &             & & \\
                            & \ordataset                & 4D-OR                     &             &            &             & & \\
                            & 4D-OR                     & \ordataset                & \checkmark  & \checkmark & \checkmark  & & \\
                            & SUSTech1K                 & 4D-OR                     & \checkmark  & \checkmark & \checkmark  & & \\
                            & SUSTech1K                 & \ordataset                & \checkmark  & \checkmark & \checkmark  & & \\ \hline
\multirow{6}{*}{PAT \cite{ni2023part}}        & 4D-OR                     & 4D-OR                     & \checkmark  & \checkmark & \checkmark  & & \\
                            & \ordataset                & \ordataset                & \checkmark  & \checkmark & \checkmark  & & \\
                            & \ordataset                & 4D-OR                     & \checkmark  & \checkmark & \checkmark  & & \\
                            & 4D-OR                     & \ordataset                & \checkmark  & \checkmark & \checkmark  & & \\
                            & SUSTech1K                 & 4D-OR                     & \checkmark  & \checkmark & \checkmark  & & \\
                            & SUSTech1K                 & \ordataset                & \checkmark  & \checkmark & \checkmark  & & \\ \hline
\multirow{6}{*}{LidarGait \cite{lidargait}}  & 4D-OR                     & 4D-OR                     &             &            &            &\checkmark & \checkmark \\
                            & \ordataset & \ordataset &             &            &            &  \checkmark &  \checkmark \\
                            & \ordataset & 4D-OR                     &             &            &            & &  \\
                            & 4D-OR                     & \ordataset &             &            &             \\
                            & SUSTech1K                 & 4D-OR                     &             &            &            & \checkmark & \checkmark  \\
                            & SUSTech1K                 & \ordataset &             &            &      & \checkmark & \checkmark   \\ \hline
\multirow{6}{*}{Ours (PC)}  & 4D-OR                     & 4D-OR                     &             &            &             & & \\
                            & \ordataset & \ordataset &             &            &            & &  \\
                            & \ordataset & 4D-OR                     &             &            &             & & \\
                            & 4D-OR                     & \ordataset &             &            &            & &  \\
                            & SUSTech1K                 & 4D-OR                     &             &            &            & &  \\
                            & SUSTech1K                 & \ordataset &             &            &   & \checkmark & \checkmark  \\
                            \bottomrule
                            
\end{tabular}
}
\end{table*}

\end{document}